\documentclass{article}
\usepackage{iclr2022_osc_workshop,times}

\usepackage{amsmath,amsfonts,bm}

\def\eqref#1{equation~\ref{#1}}
\def\1{\bm{1}}

\DeclareMathAlphabet{\mathsfit}{\encodingdefault}{\sfdefault}{m}{sl}
\SetMathAlphabet{\mathsfit}{bold}{\encodingdefault}{\sfdefault}{bx}{n}

\usepackage{hyperref}
\usepackage{url}
\usepackage{graphicx}
\usepackage{subcaption}
\usepackage{wrapfig}
\usepackage{adjustbox}
\newcommand\csize{0.07}

\usepackage{array}
\usepackage{booktabs}
\usepackage{multirow}

\usepackage{amsmath,amsfonts,amssymb,amsthm}
\usepackage{mathtools}
\usepackage{commath}

\usepackage{xcolor}
\definecolor{lightgrey}{rgb}{0.43,0.43,0.43}

\usepackage{soul}

\usepackage{enumitem}  % allows \begin{itemize}[leftmargin=*]

\definecolor{mydarkblue}{rgb}{0,0.08,0.55}
\hypersetup{  % Change citation and link colors from eye-cancer-green to nice blue.
    colorlinks=true,
    linkcolor=mydarkblue,
    citecolor=mydarkblue,
    filecolor=mydarkblue,
    urlcolor=mydarkblue
}

\title{Binding Actions to Objects in World Models}

\author{Ondrej Biza $^1$, Robert Platt $^{1,*}$, Jan-Willem van de Meent $^{1,2,*}$, \\
\textbf{Lawson L. S. Wong $^{1,*}$ and Thomas Kipf $^3$} \\
$^1$ Northeastern University, Boston, MA, USA \\
$^2$ University of Amsterdam, Netherlands \\
$^3$ Google Research, Brain Team \\
$^*$ Equal contribution. Correspondence to: \texttt{biza.o@northeastern.edu}.
}

\iclrfinalcopy
\begin{document}

\maketitle

\begin{abstract}
We study the problem of binding actions to objects in object-factored world models using action-attention mechanisms. We propose two attention mechanisms for binding actions to objects, soft attention and hard attention, which we evaluate in the context of structured world models for five environments. Our experiments show that hard attention helps contrastively-trained structured world models to learn to separate individual objects in an object-based grid-world environment. Further, we show that soft attention increases performance of factored world models trained on a robotic manipulation task. The learned action attention weights can be used to interpret the factored world model as the attention focuses on the manipulated object in the environment. Source code: \href{https://github.com/ondrejba/action_attention_iclr_22}{https://github.com/ondrejba/action\_attention\_iclr\_22}.
\end{abstract}

% keywords: world models, attention, binding, state factorization, atari, robotic manipulation
% TLDR: We present two attention mechanisms for binding actions to objects that help world models to factor states and to learn the physics of robotic pick-and-place.

\section{Introduction}

% RL often involves objects, actions usually affect a small subset of all objects
Reinforcement learning agents often operate in domains that contain multiple objects, such as robotic manipulation (e.g. \citet{janner19reasoning,li20towards}) or game playing (e.g. \citet{bellemare13arcade,guss2019minerl}). An action performed by an agent in an object-based environment is unlikely to affect all objects at once. A household robot might pick up a single object or push several objects, but it will not move all objects in a room at once. In Atari Pong, the agent only controls the paddle while the other two objects--the ball and the other player--are not directly influenced by actions. The sparsity of interactions between actions and objects is a useful inductive bias for an agent's world model.

We implement a \textit{learned} binding of an action to objects, which is constrained to be sparse. The sparsity is enforced by an attention mechanism over objects, with individual weights being constrained to sum to one using the softmax operator. Prior work considered structured action spaces in world models for grid-worlds with objects \citep{kipf20contrastive}, a Tetris-like block stacking task \citep{janner19reasoning,veerapaneni19entity} and in model-free learning for 2D construction from blocks \citep{bapst19structured}. Unlike our work, the binding of an action to a particular object, or an edge between objects in a graph, was hand-coded, without any learning. \citet{goyal21neural} proposed an attention mechanism for binding production rules \citep{llovett05thiking}, which include information about action selection, to slots in a structured model.

We propose two different attention mechanisms and test them in object-based grid worlds, two Atari games and a simulated robotic manipulation task. Firstly, we work with contrastively-learned structured world models (C-SWMs, \citet{kipf20contrastive}) that jointly learn to factor states and predict the dynamics of the environment. This class of models often requires a strong bias in order to learn to properly distinguish objects without explicit supervision. To this end, we encode the bias that an action affects exactly one object using a categorical random variable. We call this approach \textit{hard attention}, drawing on computer vision literature \citep{sharma15action}. Secondly, we test our idea in a realistic robotic manipulation domain, where the individual states are already factored into object and our task is to learn factored world models (FWMs, \citet{biza2022factored}). Here, the robot is tasked with building towers from blocks; all blocks in a tower might potentially be affected by a robot's action. Hence, the categorical action binding bias is too restrictive. Instead, we propose a \textit{soft attention} module, which transforms and weights actions for each object based on their predicted impact on said object. It is based on single-head self-attention \citep{vaswani17attention}. We limit our exploration of action binding to C-SWMs and FWMs; see \citet{battaglia18relational,greff20binding} for a review of other structured model architectures.

Our results indicate two positive outcomes and one negative outcome. When each object can be individually affected by an action, hard action attention proves an excellent inductive bias for learning to factor states. Conversely, when some objects are not directly affected by actions, neither soft nor hard action attention makes a difference. Finally, we show that soft attention can lead to large improvements in a pick-and-place robotic manipulation task.

In summary, our main contributions are: \\
\textbf{\#1:} We propose two novel attention modules over actions, soft and hard attention, that improve the ability of structured models to factor states and to predict transition dynamics.  \\
\textbf{\#2:} We identify the phenomenon wherein a structured world model captures highly correlated information in all its available slots. This happens in Atari Pong and Space Invaders, where only the player-controlled paddle or spaceship is directly affected by actions.

\section{Preliminaries: Structured and Factored World Models}

World models are trained to make predictions about the state of an environment $s^0$ after applying a sequence of actions $a^1, a^2,$ ... $a^T$. It is commonly assumed that the state is represented as an image and the transition dynamics are deterministic. We make no assumptions about the action space. World models used in this paper learn a latent space without image reconstruction. That is, the model encodes $s^0$ into a latent encoding $z^0$ and predicts the future $z^{1:T}$ based on $a^{1:T}$.

\textbf{Structured world models} encode an image $s$ into a latent state composed of K slots $z_{1:K}$, each represented as a feature vector. The model learns to assign objects in $s$ to individual slots (i.e. to factor the scene) without any direct supervision. We use the contrastively-trained structured world model (C-SWM, \citet{kipf20contrastive})--it learns structured embeddings using a convolutional encoder $E_{\phi}$ and predicts the effects of actions using a graph neural network (GNN) $T_{\theta}$ \citep{gori05new}. The GNN uses a $f_{\text{node}}$ and an edge network $f_{\text{edge}}$ to compute the next latent state of the $k$th object:
\begin{align}
    \hat{z}^{t+1}_k = z^t_k + T_{\theta}(z^t, a^t)_k = z^t_k + f_{\text{node}}\bigg(z^t_k, a^t, \sum_{i \neq k} f_{\text{edge}}(z^t_i, z^t_k)\bigg).
\end{align}
The model is trained by contrastive learning using the following loss:
\begin{align}
    L(z^t, \bar{z}, z^{t+1}, \hat{z}^{t+1}) = \sum_{k} \norm{z^{t+1}_k - \hat{z}^{t+1}_k}_2^2 - \max\bigg\{0, \gamma - \sum_{k}\norm{z^t_k - \bar{z}_k}_2^2\bigg\}.
\end{align}
The negative example $\bar{z}$ is an encoding of a state randomly sampled form the training set and $\gamma$ is a hyper-parameter. Intuitively, the negative term in the loss states that a pair of randomly sampled states should be at least $\gamma$ apart in the embedding space. This prevents the model from converging to a trivial solution.

We use \textbf{factored world models} to refer to world models that operate over factored states. That is, they learn to model the transition dynamics of individual factors without having to learn to discover factors. \citet{biza2022factored} proposed a factored world model for pick-and-place robotic manipulation tasks. The model receives a factored state $s_{1:K}$ composed of images of the $K$ object present in a scene. Each state factor is individually encoded into a latent factor: $z_{1:K} = \langle f_{\text{enc}}(s_1), f_{\text{enc}}(s_2), ..., f_{\text{enc}}(s_K) \rangle$. The transition model and the training loss follow C-SWM with the exception of two design decisions in the transition model. (1) several layers of GNNs are used and skip connections are added in between layers and (2) the action is provided not only to the node network $f_{node}$ but also to the edge network $f_{edge}$. We experiment with different numbers of GNN layers, but omit the edge actions.

\section{Binding Actions to Objects}

In this section, we describe two attention mechanisms that bind actions to object slots. In both cases, we start with a factored latent state $z_{1:K}$, which is an output of an encoder of either C-SWM or FWM. We want to compare the individual factors to the action in order to determine which slots are affected. First, we transform individual factors into key vectors, $k = \langle j_k(z_1), j_k(z_2), ..., j_k(z_K) \rangle$, and the action into a query vector $q = j_q(a)$. Both $j_k$ and $j_q$ are Multi-Layer Perceptrons. The keys and queries are used to compute attention weights $\alpha$ over object slots $1$ to $K$ for action $a$:
\begin{align}
    \alpha &= \text{softmax}(k_1^T q, k_2^T q, ..., k_K^T q).
\end{align}
It is important to note that in both the hard and soft attention modules described below, the attention weights end up transforming the action that is given to the node network of the GNN transition model. For the $k$th object slot, instead of computing $f_{\text{node}}(z_k, a, ...)$ we will use $f_{\text{node}}(z_k, a'_k, ...)$, where $a'_k$ is an action constructed specifically for the $k$th object slot.

\textbf{Hard attention:} We define a categorical variable $M$ that represents the membership of an action to a particular factor:
\begin{align}
    M \sim \text{Categorical}(\alpha_1, \alpha_2, ..., \alpha_K).
\end{align}
This random variable only plays a role when predicting the transitions and computing the positive term of the contrastive loss (the negative term only depends on the current encoded state). During training, we compute the expectation of $M$ via summation over all of its possible assignments (same as the number of object slots / factors). During inference, we take the assignment with the highest probability. The random variable is used as follows:
\begin{align}
    \hat{z}^{t+1}_k = z^t_k + T_{\theta}(z^t, \text{pad}(a^t, m))_k \quad | \quad M = m,
\end{align}
with the pad$(a^t, m)$ operator assigning action $a^t$ to the $m$th slot and zeros to all other slots. Finally, the positive term of the contrastive loss function is computed as follows:
\begin{align}
    \mathbb{E}_M \left[ \norm{z^{t+1}_k - \hat{z}^{t+1}_k}_2^2 \right] = \sum_{m=1}^K \alpha_m \norm{ z^{t+1}_k - \left[ z^t_k + T_{\theta}(z^t, \text{pad}(a^t, m))_k \right] }_2^2.
\end{align}

\textbf{Soft attention:} Unlike hard attention, soft attention weights individual actions provided to slots by attention weights without defining a probabilistic model. We follow the single-head self-attention module from the Transformer \citep{vaswani17attention}. The action $a$ is first transformed into a value vector $v = j_v(a)$. Then, the value vector is multiplied by the previously computed attention weights for each factor: $a' = \langle \alpha_1 v, \alpha_2 v, ..., \alpha_K v \rangle$. We then pass the (newly) factored action $a'$ into the transition model. Note the individual factors in the action are all equal up to a multiplicative factor. Table \ref{tab:fwm_attention} in the Appendix visualizes $\alpha$ in a robotic pick-and-place task.

\section{Experiments}

\begin{table*}[t!]
\centering
\caption{\label{tab:cswm} Results for 2D grid-world \texttt{Shapes} and 3D grid-world \texttt{Cubes} with unfactored states and actions (different from \citet{kipf20contrastive}, where actions are factored) as well as Atari Pong and Space Invader datasets from \citet{biza21impact}. We report means and standard deviations for Hits@1 and Mean Reciprocal Rank for 1, 5 and 10 step predictions. Each model was run 20 times in \texttt{Shapes} and \texttt{Cubes}, and 4 times in Atari. The results for Atari differ from \citet{biza21impact} because we use 1k instead of a 100 negative states during evaluation, making the task harder. Additionally, we report correlation between individual slots of C-SWM.}

\begin{adjustbox}{center,width=0.9\linewidth}
\begin{tabular}{lllrrrrrrr}
\toprule
  & & & \multicolumn{2}{c}{1 Step} & \multicolumn{2}{c}{5 Steps} & \multicolumn{2}{c}{10 Steps} & Slot \\ \cmidrule(lr){4-5} \cmidrule(lr){6-7} \cmidrule(lr){8-9}
&&Model & H@1 & MRR & H@1 & MRR & H@1 & MRR & Correl. \\ \midrule
\parbox[t]{0.1mm}{\multirow{3}{*}{\rotatebox[origin=c]{45}{2D GRID}}} & \parbox[t]{9mm}{\multirow{3}{*}{\rotatebox[origin=c]{45}{SHAPES}}} & C-SWM & {28.7}{\color{lightgrey}\tiny$\pm$2.6} & {42.9}{\color{lightgrey}\tiny$\pm$2.6} & {3.7}{\color{lightgrey}\tiny$\pm$0.8} & {10.9}{\color{lightgrey}\tiny$\pm$1.6} & {1.5}{\color{lightgrey}\tiny$\pm$0.4} & {5.4}{\color{lightgrey}\tiny$\pm$1.0} & {1.00}{\color{lightgrey}\tiny$\pm$0.00} \\
&& Soft Attention & {30.9}{\color{lightgrey}\tiny$\pm$2.9} & {45.7}{\color{lightgrey}\tiny$\pm$2.6} & {4.3}{\color{lightgrey}\tiny$\pm$0.8} & {12.5}{\color{lightgrey}\tiny$\pm$1.5} & {1.7}{\color{lightgrey}\tiny$\pm$0.4} & {6.3}{\color{lightgrey}\tiny$\pm$1.0} & {1.00}{\color{lightgrey}\tiny$\pm$0.00} \\
&& {Hard Attention} & {\bf 98.7}{\color{lightgrey}\tiny$\pm$5.3} & {\bf 99.1}{\color{lightgrey}\tiny$\pm$3.6} & {\bf 95.5}{\color{lightgrey}\tiny$\pm$16.9} & {\bf 96.3}{\color{lightgrey}\tiny$\pm$14.5} & {\bf 93.4}{\color{lightgrey}\tiny$\pm$20.5} & {\bf 94.3}{\color{lightgrey}\tiny$\pm$18.7} & {0.08}{\color{lightgrey}\tiny$\pm$0.08} \\
\midrule
\parbox[t]{0.1mm}{\multirow{3}{*}{\rotatebox[origin=c]{45}{3D GRID}}} & \parbox[t]{9mm}{\multirow{3}{*}{\rotatebox[origin=c]{45}{CUBES\ \ }}} & C-SWM & {32.0}{\color{lightgrey}\tiny$\pm$3.4} & {46.3}{\color{lightgrey}\tiny$\pm$3.1} & {4.8}{\color{lightgrey}\tiny$\pm$1.2} & {13.3}{\color{lightgrey}\tiny$\pm$2.1} & {2.0}{\color{lightgrey}\tiny$\pm$0.5} & {7.0}{\color{lightgrey}\tiny$\pm$1.4} & {1.00}{\color{lightgrey}\tiny$\pm$0.00} \\
&& Soft Attention & {40.8}{\color{lightgrey}\tiny$\pm$12.1} & {54.5}{\color{lightgrey}\tiny$\pm$10.6} & {9.0}{\color{lightgrey}\tiny$\pm$6.3} & {20.0}{\color{lightgrey}\tiny$\pm$8.4} & {4.0}{\color{lightgrey}\tiny$\pm$2.9} & {11.3}{\color{lightgrey}\tiny$\pm$5.1} & {0.99}{\color{lightgrey}\tiny$\pm$0.02} \\
&& {Hard Attention} & {\bf 97.3}{\color{lightgrey}\tiny$\pm$5.9} & {\bf 98.3}{\color{lightgrey}\tiny$\pm$3.9} & {\bf 88.8}{\color{lightgrey}\tiny$\pm$18.2} & {\bf 91.6}{\color{lightgrey}\tiny$\pm$14.3} & {\bf 79.2}{\color{lightgrey}\tiny$\pm$24.5} & {\bf 83.3}{\color{lightgrey}\tiny$\pm$20.8} & {0.21}{\color{lightgrey}\tiny$\pm$0.18} \\
\midrule
\parbox[t]{0.1mm}{\multirow{3}{*}{\rotatebox[origin=c]{45}{ATARI}}} & \parbox[t]{9mm}{\multirow{3}{*}{\rotatebox[origin=c]{45}{PONG}}} & C-SWM & {93.2}{\color{lightgrey}\tiny$\pm$1.1} & {96.3}{\color{lightgrey}\tiny$\pm$0.6} & {\bf 36.3}{\color{lightgrey}\tiny$\pm$3.8} & {\bf 49.5}{\color{lightgrey}\tiny$\pm$3.8} & {\bf 11.2}{\color{lightgrey}\tiny$\pm$2.4} & {\bf 19.5}{\color{lightgrey}\tiny$\pm$1.8} & {0.84}{\color{lightgrey}\tiny$\pm$0.11} \\
&& {Soft Attention} & {\bf 93.8}{\color{lightgrey}\tiny$\pm$0.8} & {\bf 96.6}{\color{lightgrey}\tiny$\pm$0.4} & {26.1}{\color{lightgrey}\tiny$\pm$6.3} & {38.5}{\color{lightgrey}\tiny$\pm$6.5} & {7.3}{\color{lightgrey}\tiny$\pm$4.3} & {13.0}{\color{lightgrey}\tiny$\pm$5.6} & {0.87}{\color{lightgrey}\tiny$\pm$0.10}  \\
&& {Hard Attention} & {92.9}{\color{lightgrey}\tiny$\pm$0.2} & {96.2}{\color{lightgrey}\tiny$\pm$0.1} & {28.3}{\color{lightgrey}\tiny$\pm$7.1} & {41.9}{\color{lightgrey}\tiny$\pm$6.9} & {9.6}{\color{lightgrey}\tiny$\pm$3.6} & {16.8}{\color{lightgrey}\tiny$\pm$4.7} & {0.85}{\color{lightgrey}\tiny$\pm$0.10} \\
\midrule
\parbox[t]{0.1mm}{\multirow{3}{*}{\rotatebox[origin=c]{45}{ATARI}}} & \parbox[t]{9mm}{\multirow{3}{*}{\rotatebox[origin=c]{45}{SPACE}}} & C-SWM & {\bf 97.1}{\color{lightgrey}\tiny$\pm$0.2} & {\bf 98.5}{\color{lightgrey}\tiny$\pm$0.1} & {86.0}{\color{lightgrey}\tiny$\pm$0.9} & {\bf 91.7}{\color{lightgrey}\tiny$\pm$0.5} & {74.9}{\color{lightgrey}\tiny$\pm$1.7} & {83.4}{\color{lightgrey}\tiny$\pm$1.0} & {0.82}{\color{lightgrey}\tiny$\pm$0.13} \\
&& Soft Attention & {96.7}{\color{lightgrey}\tiny$\pm$0.6} & {98.3}{\color{lightgrey}\tiny$\pm$0.3} & {\bf 86.2}{\color{lightgrey}\tiny$\pm$0.8} & {\bf 91.7}{\color{lightgrey}\tiny$\pm$0.5} & {\bf 76.2}{\color{lightgrey}\tiny$\pm$0.7} & {\bf 84.0}{\color{lightgrey}\tiny$\pm$0.3} & {0.67}{\color{lightgrey}\tiny$\pm$0.05} \\
&& {Hard Attention} & {96.8}{\color{lightgrey}\tiny$\pm$0.7} & {98.3}{\color{lightgrey}\tiny$\pm$0.3} & {84.0}{\color{lightgrey}\tiny$\pm$1.3} & {90.2}{\color{lightgrey}\tiny$\pm$0.8} & {73.9}{\color{lightgrey}\tiny$\pm$2.0} & {82.3}{\color{lightgrey}\tiny$\pm$1.2} & {0.80}{\color{lightgrey}\tiny$\pm$0.13} \\
\bottomrule
\end{tabular}
\end{adjustbox}
\end{table*}

We aim to answer the following questions:
\begin{enumerate}[leftmargin=*]
    \item Does hard attention help C-SWM factor environments where each object can be independently affected by an action (\texttt{2D Shapes} and \texttt{3D Cubes})?
    \item Does the same apply when some objects are only indirectly (or not at all) affected by actions (Atari Pong and Space Invaders)?
    \item Does soft attention increase the performance of factored world models in robotic manipulation?
\end{enumerate}

\subsection{Environments}

We use two grid-world (\texttt{2D Shapes} and \texttt{3D Cubes}) and two Atari environments (Atari Pong and Space Inveders) from \citet{kipf20contrastive} and a robotic manipulation environment from \citet{biza2022factored}. We make changes to \texttt{2D Shapes} and \texttt{3D Cubes} in order to better study the effect of action attention. We describe each environment below.

\begin{itemize}[leftmargin=*]
    \item \textbf{2D Shapes and 3D Cubes:} In both environments, five objects of different shapes and colors are placed in a $5{\times}5$ grid-world. Each object can be moved independently and the objects are not allowed to pass through each other. Crucially, \citet{kipf20contrastive} represented actions as a pair of object index and direction of movement. As the action space was factored by design, C-SWM could use this information to make state factorization easier. We change the action space to represent object position and direction of movement. Hence, the model does not know the index of the object being moved, making state factorization more challenging.
    \item \textbf{Atari Pong and Space Invaders}: C-SWM is trained and tested on modeling short action sequences (10 timesteps) in these two games. We use the approach from \citet{biza21impact}, where a trained agent acts for a random number of steps to reach an interesting starting state from which a random policy is rolled out. The collected datasets contain only the random rollouts so that the agent experiences both good and bad actions.
    \item \textbf{Robotic Pick-and-Place:} A simulated UR5 robotic arm manipulates six cubes. The action space is the continuous space of $(x,y)$ coordinates, where a pre-programmed \texttt{pick} or \texttt{place} action is executed. The environment is observed by two side-viewing RGB cameras and additional two RGB cameras capture the content of the robot's hand. These observations are post-processed to extract individual objects and their bounding boxes to capture each object independently (i.e. a factored state space). The training task is to make a tower of four cubes; the model is then expected to transfer to four testing tasks--wall, stairs, two towers of three and three towers of two--without additional training. 
\end{itemize}

\subsection{Shapes, Cubes and Atari Games}

\textbf{Setup:} The C-SWMs are evaluated on their ability to predict future latent states in an evaluation dataset using ranking metrics (Hits@1 and MRR, see Appendix \ref{ap:metrics}). These metrics check if the predicted latent state is closer to the encoding of the true next state compared to a set of negative states. Intuitively, we both test the ability of the model to predict the future and the ability of the encoder to distinguish between states.

\textbf{Result:} Ranking scores are reported in Table \ref{tab:cswm}. We would like to highlight the performance of hard action attention in \texttt{2D Shapes} and \texttt{3D Cubes}. In these environments, baseline C-SWM fails because the action space is not factored per individual object. In contrast, C-SWM with hard attention encodes the correct inductive bias and learns to assign individual objects to separate slots in most cases. 

We do not see the same increase in performance in Atari games: baseline C-SWM, soft attention and hard attention achieve comparable results. To explain this result, we introduce a new metric to measure the correlation between individual slots. If all slots capture the same information given a single input state, the correlation coefficient is going to be 1 (please see Appendix \ref{ap:metrics}). We see that successful models in \texttt{2D Shapes} and \texttt{3D Cubes} have a low slot correlation, as each individual object should be represented by a different slot. Conversely, all models in Atari (as well as failed models in \texttt{2D Shapes} and \texttt{3D Cubes}) have high correlation between slots. We show one example of this phenomenon in Figure \ref{fig:pong_correlation} in the Appendix where the latent space of all slots is nearly identical (the first slot seems to be flipped, but follows the same pattern).

The correlation metric results suggest that C-SWM has no need for more than one slot in Atari games. This is supported by the original results from \citet{kipf20contrastive}. We hypothesize that since the action often affects only one object (the paddle in Pong and the spaceship in Space Invaders), there is not enough information to guide C-SWM to factor objects not directly affected by the action.

\subsection{Robotic Pick-and-Place}

\begin{figure}[t!]
    \begin{subfigure}{0.33\textwidth}
        \centering
        \includegraphics[width=1\textwidth]{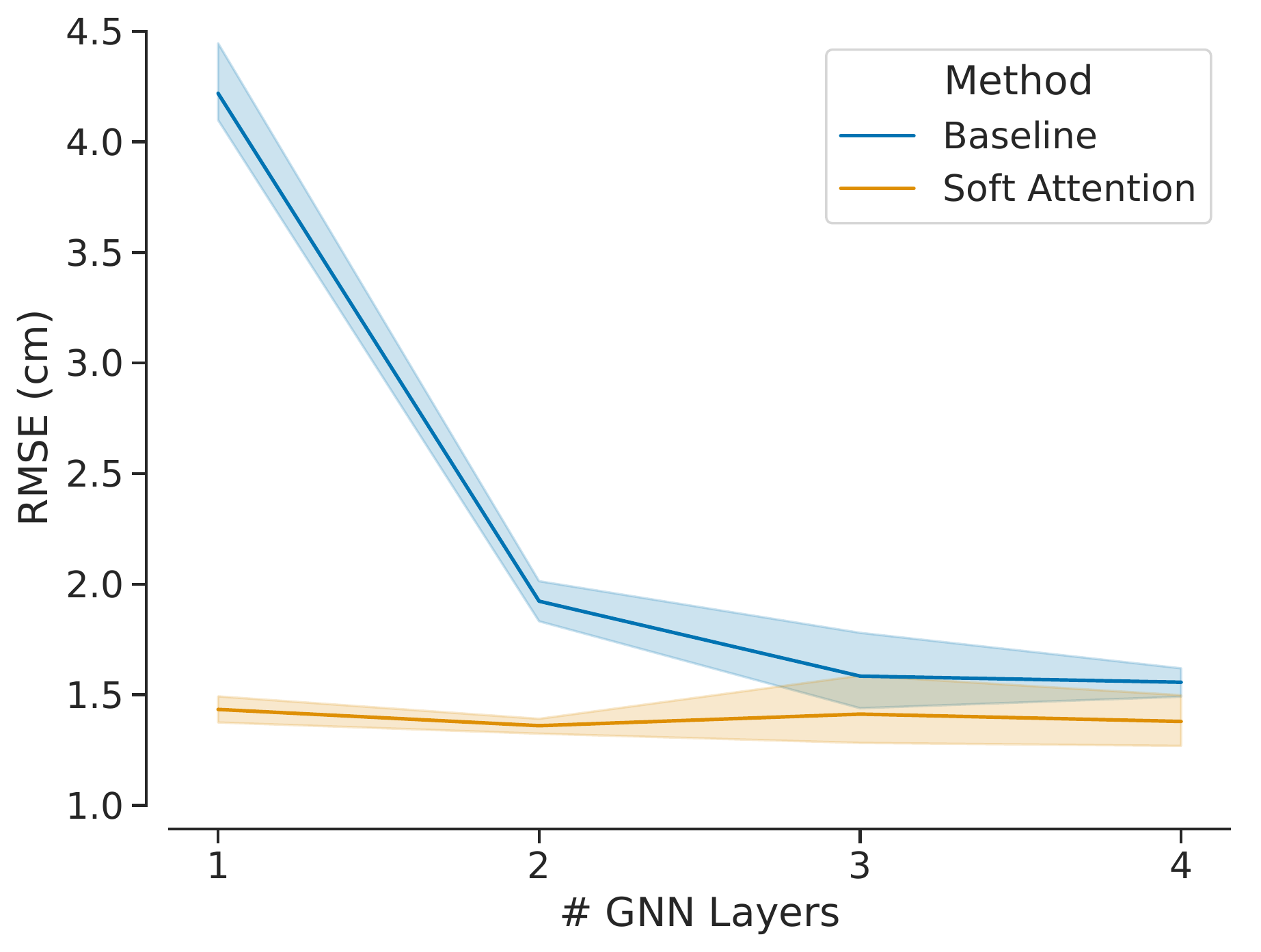}
    \end{subfigure}
    \begin{subfigure}{0.33\textwidth}
        \centering
        \includegraphics[width=1\textwidth]{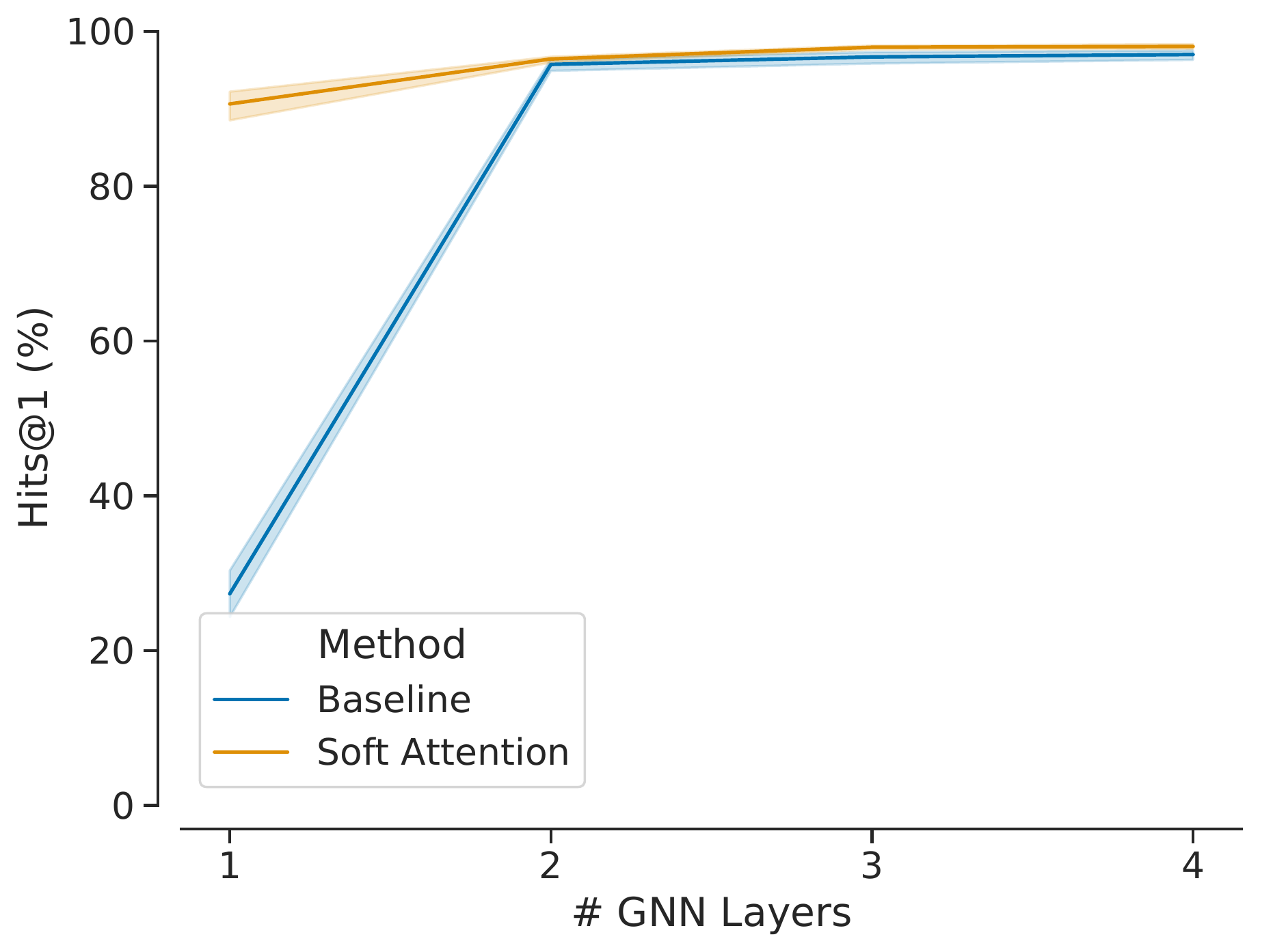}
    \end{subfigure}
    \begin{subfigure}{0.33\textwidth}
        \centering
        \vspace{-1em}
        \includegraphics[width=0.66\textwidth]{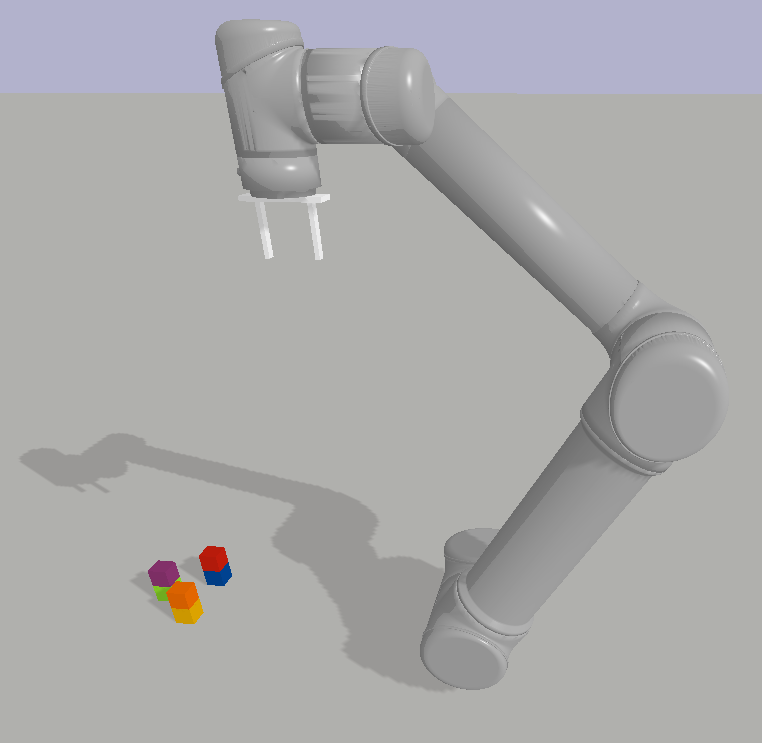}
    \end{subfigure}
    \caption{Zero-shot transfer results in block position prediction error (left, the lower the better) and action action sequence ranking (middle, the higher the better) in a UR5 pick-and-place environment with six cubes (right). We average over four random seeds and shade the 95\% confidence interval.}
    \label{fig:fwm}
\end{figure}

\textbf{Setup:} We replicate the experiment from \citet{biza2022factored}, where a UR5 robotic arm picks and places six cubes. A dataset of expert demonstrations is available for the training task and the model is evaluated on its ability to predict the block positions and rank action sequences in expert demonstrations of the zero-shot transfer tasks. The block position prediction metric measures if ground-truth information about the positions of all blocks can be decoded from the learned latent space. The action ranking metric measures if the model can distinguish action sequences that successfully solve a task from perturbed action sequences that do not. See Appendix \ref{ap:metrics} for additional information.

\textbf{Result:} Factored world models with soft action attention outperform the baseline models when the number of GNN layers is set to one or two (Figure \ref{fig:fwm}). As the number of layers increases to three and four, the performance of the two models converges. The attention weights predicted by Soft Attention for a model with one graph neural network layer are shown in Table \ref{tab:fwm_attention} in the Appendix. The model correctly identifies which object is being manipulated. Additionally, the model pays attention to where a particular cube is being placed. When building a tower, the cubes below the placed object also receive the information about the \texttt{place} action with a low (but non-zero) weight. We believe this behavior emerged because placement of a cube can often move the cubes below.

Furthermore, we hypothesize that the diminishing benefit of soft attention as we add more GNN layers can be explained by the increasing capacity of the model. A single GNN layer can benefit from the pre-computed information about which object will be affected. On the other hand, a stack of four GNN layers receives the information about the action at each layer, making it easier to distinguish affected and unaffected objects within the transition model.

\section{Conclusion}

We propose two attention mechanisms for binding actions to objects, soft attention and hard attention, and validate them on the task of learning structured world models in five environments. As one of our reviewers pointed out, "there seems to be a dilemma between soft and hard attention -- the soft attention is more expressive, however, it also under-performs hard attention in some experiments." We view this trade-off from the perspective of expressiveness versus bias. Hard attention provides a strong bias (that all actions move exactly one object), which allows C-SWM to converge to the right solution in the grid-world environments. Conversely, soft attention is more flexible in providing information about the action to several objects. This construction fits our simulated robotic manipulation task, where several cubes might move at once when they are being stacked. 

\section*{Acknowledgements}

This work is supported in part by NSF 1724257, NSF 1724191, NSF 1763878, NSF 1750649, NSF 1835309, NSF 2107256 and
NASA 80NSSC19K1474.

\bibliography{iclr2022_osc_workshop}

\begin{thebibliography}{16}
\providecommand{\natexlab}[1]{#1}
\providecommand{\url}[1]{\texttt{#1}}
\expandafter\ifx\csname urlstyle\endcsname\relax
  \providecommand{\doi}[1]{doi: #1}\else
  \providecommand{\doi}{doi: \begingroup \urlstyle{rm}\Url}\fi

\bibitem[Bapst et~al.(2019)Bapst, Sanchez{-}Gonzalez, Doersch, Stachenfeld,
  Kohli, Battaglia, and Hamrick]{bapst19structured}
Victor Bapst, Alvaro Sanchez{-}Gonzalez, Carl Doersch, Kimberly~L. Stachenfeld,
  Pushmeet Kohli, Peter~W. Battaglia, and Jessica~B. Hamrick.
\newblock Structured agents for physical construction.
\newblock In Kamalika Chaudhuri and Ruslan Salakhutdinov (eds.),
  \emph{Proceedings of the 36th International Conference on Machine Learning,
  {ICML} 2019, 9-15 June 2019, Long Beach, California, {USA}}, volume~97 of
  \emph{Proceedings of Machine Learning Research}, pp.\  464--474. {PMLR},
  2019.

\bibitem[Battaglia et~al.(2018)Battaglia, Hamrick, Bapst, Sanchez{-}Gonzalez,
  Zambaldi, Malinowski, Tacchetti, Raposo, Santoro, Faulkner,
  G{\"{u}}l{\c{c}}ehre, Song, Ballard, Gilmer, Dahl, Vaswani, Allen, Nash,
  Langston, Dyer, Heess, Wierstra, Kohli, Botvinick, Vinyals, Li, and
  Pascanu]{battaglia18relational}
Peter~W. Battaglia, Jessica~B. Hamrick, Victor Bapst, Alvaro
  Sanchez{-}Gonzalez, Vin{\'{\i}}cius~Flores Zambaldi, Mateusz Malinowski,
  Andrea Tacchetti, David Raposo, Adam Santoro, Ryan Faulkner, {\c{C}}aglar
  G{\"{u}}l{\c{c}}ehre, H.~Francis Song, Andrew~J. Ballard, Justin Gilmer,
  George~E. Dahl, Ashish Vaswani, Kelsey~R. Allen, Charles Nash, Victoria
  Langston, Chris Dyer, Nicolas Heess, Daan Wierstra, Pushmeet Kohli, Matthew
  Botvinick, Oriol Vinyals, Yujia Li, and Razvan Pascanu.
\newblock Relational inductive biases, deep learning, and graph networks.
\newblock \emph{CoRR}, abs/1806.01261, 2018.

\bibitem[Bellemare et~al.(2013)Bellemare, Naddaf, Veness, and
  Bowling]{bellemare13arcade}
Marc~G. Bellemare, Yavar Naddaf, Joel Veness, and Michael Bowling.
\newblock The arcade learning environment: An evaluation platform for general
  agents.
\newblock \emph{J. Artif. Intell. Res.}, 47:\penalty0 253--279, 2013.

\bibitem[Biza et~al.(2021)Biza, van~der Pol, and Kipf]{biza21impact}
Ondrej Biza, Elise van~der Pol, and Thomas Kipf.
\newblock The impact of negative sampling on contrastive structured world
  models.
\newblock \emph{ICML 2021 Workshop: Self-Supervised Learning for Reasoning and
  Perception}, 2021.

\bibitem[Biza et~al.(2022)Biza, Kipf, Klee, Platt, van~de Meent, and
  Wong]{biza2022factored}
Ondrej Biza, Thomas Kipf, David Klee, Robert Platt, Jan-Willem van~de Meent,
  and Lawson L.~S. Wong.
\newblock Factored world models for zero-shot generalization in robotic
  manipulation, 2022.

\bibitem[Gori et~al.(2005)Gori, Monfardini, and Scarselli]{gori05new}
M.~Gori, G.~Monfardini, and F.~Scarselli.
\newblock A new model for learning in graph domains.
\newblock In \emph{Proceedings. 2005 IEEE International Joint Conference on
  Neural Networks, 2005.}, volume~2, pp.\  729--734 vol. 2, 2005.

\bibitem[Goyal et~al.(2021)Goyal, Didolkar, Ke, Blundell, Beaudoin, Heess,
  Mozer, and Bengio]{goyal21neural}
Anirudh Goyal, Aniket~Rajiv Didolkar, Nan~Rosemary Ke, Charles Blundell,
  Philippe Beaudoin, Nicolas Heess, Michael~Curtis Mozer, and Yoshua Bengio.
\newblock Neural production systems.
\newblock In A.~Beygelzimer, Y.~Dauphin, P.~Liang, and J.~Wortman Vaughan
  (eds.), \emph{Advances in Neural Information Processing Systems}, 2021.

\bibitem[Greff et~al.(2020)Greff, van Steenkiste, and
  Schmidhuber]{greff20binding}
Klaus Greff, Sjoerd van Steenkiste, and J{\"{u}}rgen Schmidhuber.
\newblock On the binding problem in artificial neural networks.
\newblock \emph{CoRR}, abs/2012.05208, 2020.

\bibitem[Guss et~al.(2019)Guss, Houghton, Topin, Wang, Codel, Veloso, and
  Salakhutdinov]{guss2019minerl}
William~H Guss, Brandon Houghton, Nicholay Topin, Phillip Wang, Cayden Codel,
  Manuela Veloso, and Ruslan Salakhutdinov.
\newblock Minerl: A large-scale dataset of minecraft demonstrations.
\newblock \emph{arXiv preprint arXiv:1907.13440}, 2019.

\bibitem[Janner et~al.(2019)Janner, Levine, Freeman, Tenenbaum, Finn, and
  Wu]{janner19reasoning}
Michael Janner, Sergey Levine, William~T. Freeman, Joshua~B. Tenenbaum, Chelsea
  Finn, and Jiajun Wu.
\newblock Reasoning about physical interactions with object-oriented prediction
  and planning.
\newblock In \emph{7th International Conference on Learning Representations,
  {ICLR} 2019, New Orleans, LA, USA, May 6-9, 2019}. OpenReview.net, 2019.

\bibitem[Kipf et~al.(2020)Kipf, van~der Pol, and Welling]{kipf20contrastive}
Thomas~N. Kipf, Elise van~der Pol, and Max Welling.
\newblock Contrastive learning of structured world models.
\newblock In \emph{8th International Conference on Learning Representations,
  {ICLR} 2020, Addis Ababa, Ethiopia, April 26-30, 2020}. OpenReview.net, 2020.

\bibitem[Li et~al.(2020)Li, Jabri, Darrell, and Agrawal]{li20towards}
Richard Li, Allan Jabri, Trevor Darrell, and Pulkit Agrawal.
\newblock Towards practical multi-object manipulation using relational
  reinforcement learning.
\newblock In \emph{2020 {IEEE} International Conference on Robotics and
  Automation, {ICRA} 2020, Paris, France, May 31 - August 31, 2020}, pp.\
  4051--4058. {IEEE}, 2020.

\bibitem[Lovett \& R.~Anderson(2005)Lovett and R.~Anderson]{llovett05thiking}
Marsha Lovett and John R.~Anderson.
\newblock Thinking as a production system.
\newblock \emph{The Cambridge handbook of thinking and reasoning}, 2005.

\bibitem[Sharma et~al.(2015)Sharma, Kiros, and Salakhutdinov]{sharma15action}
Shikhar Sharma, Ryan Kiros, and Ruslan Salakhutdinov.
\newblock Action recognition using visual attention.
\newblock \emph{CoRR}, abs/1511.04119, 2015.

\bibitem[Vaswani et~al.(2017)Vaswani, Shazeer, Parmar, Uszkoreit, Jones, Gomez,
  Kaiser, and Polosukhin]{vaswani17attention}
Ashish Vaswani, Noam Shazeer, Niki Parmar, Jakob Uszkoreit, Llion Jones,
  Aidan~N. Gomez, Lukasz Kaiser, and Illia Polosukhin.
\newblock Attention is all you need.
\newblock In Isabelle Guyon, Ulrike von Luxburg, Samy Bengio, Hanna~M. Wallach,
  Rob Fergus, S.~V.~N. Vishwanathan, and Roman Garnett (eds.), \emph{Advances
  in Neural Information Processing Systems 30: Annual Conference on Neural
  Information Processing Systems 2017, December 4-9, 2017, Long Beach, CA,
  {USA}}, pp.\  5998--6008, 2017.

\bibitem[Veerapaneni et~al.(2019)Veerapaneni, Co{-}Reyes, Chang, Janner, Finn,
  Wu, Tenenbaum, and Levine]{veerapaneni19entity}
Rishi Veerapaneni, John~D. Co{-}Reyes, Michael Chang, Michael Janner, Chelsea
  Finn, Jiajun Wu, Joshua~B. Tenenbaum, and Sergey Levine.
\newblock Entity abstraction in visual model-based reinforcement learning.
\newblock In Leslie~Pack Kaelbling, Danica Kragic, and Komei Sugiura (eds.),
  \emph{3rd Annual Conference on Robot Learning, CoRL 2019, Osaka, Japan,
  October 30 - November 1, 2019, Proceedings}, volume 100 of \emph{Proceedings
  of Machine Learning Research}, pp.\  1439--1456. {PMLR}, 2019.

\end{thebibliography}
\bibliographystyle{iclr2022_osc_workshop}

\clearpage
\appendix
\section{Appendix}

\begin{figure}[t]
    \centering
    \begin{subfigure}{\csize\textwidth}
        \centering
        \textbf{State:}
    \end{subfigure}
    \begin{subfigure}{\csize\textwidth}
        \centering
        \includegraphics[width=1\textwidth]{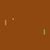}
    \end{subfigure}
    \begin{subfigure}{\csize\textwidth}
        \centering
        \includegraphics[width=1\textwidth]{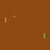}
    \end{subfigure}
    \begin{subfigure}{\csize\textwidth}
        \centering
        \includegraphics[width=1\textwidth]{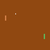}
    \end{subfigure}
    \begin{subfigure}{\csize\textwidth}
        \centering
        \includegraphics[width=1\textwidth]{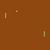}
    \end{subfigure}
    \begin{subfigure}{\csize\textwidth}
        \centering
        \includegraphics[width=1\textwidth]{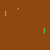}
    \end{subfigure}
    \begin{subfigure}{\csize\textwidth}
        \centering
        \includegraphics[width=1\textwidth]{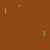}
    \end{subfigure}
    \begin{subfigure}{\csize\textwidth}
        \centering
        \includegraphics[width=1\textwidth]{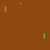}
    \end{subfigure}
    \begin{subfigure}{\csize\textwidth}
        \centering
        \includegraphics[width=1\textwidth]{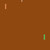}
    \end{subfigure}
    \begin{subfigure}{\csize\textwidth}
        \centering
        \includegraphics[width=1\textwidth]{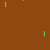}
    \end{subfigure}
    \begin{subfigure}{\csize\textwidth}
        \centering
        \includegraphics[width=1\textwidth]{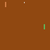}
    \end{subfigure}
    \begin{subfigure}{\csize\textwidth}
        \centering
        \includegraphics[width=1\textwidth]{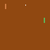}
    \end{subfigure}
    
    \begin{subfigure}{\csize\textwidth}
        \centering
        \textbf{Slot 1:}
    \end{subfigure}
    \begin{subfigure}{\csize\textwidth}
        \centering
        \includegraphics[width=1\textwidth]{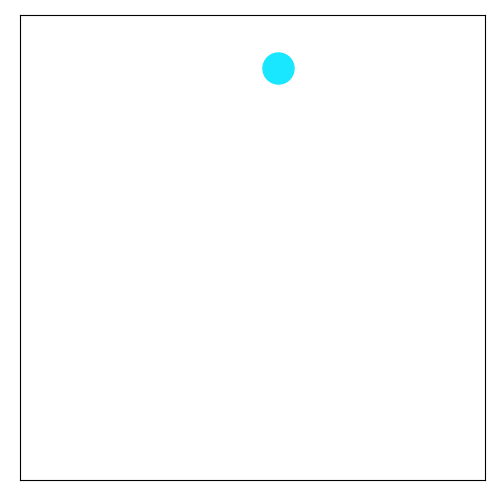}
    \end{subfigure}
        \begin{subfigure}{\csize\textwidth}
        \centering
        \includegraphics[width=1\textwidth]{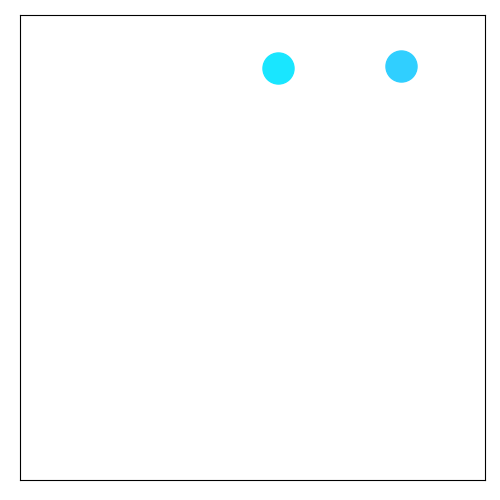}
    \end{subfigure}
        \begin{subfigure}{\csize\textwidth}
        \centering
        \includegraphics[width=1\textwidth]{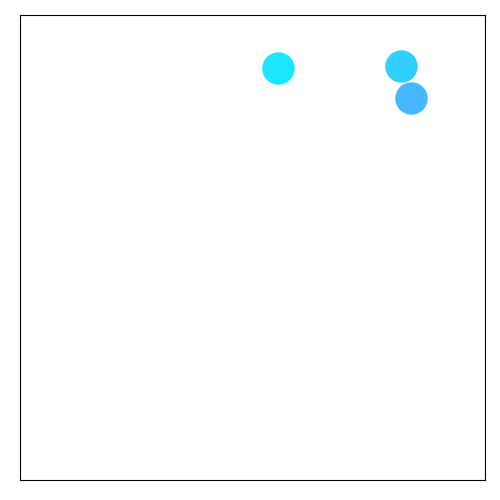}
    \end{subfigure}
        \begin{subfigure}{\csize\textwidth}
        \centering
        \includegraphics[width=1\textwidth]{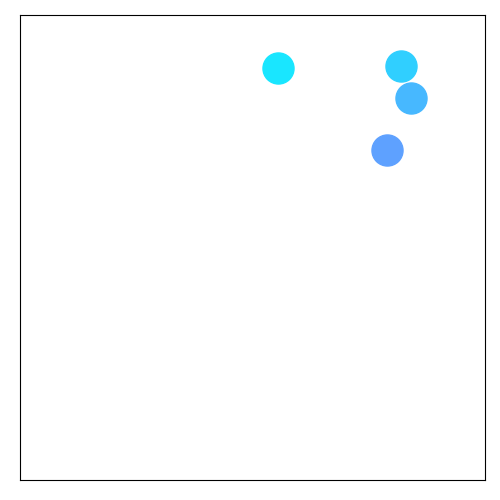}
    \end{subfigure}
        \begin{subfigure}{\csize\textwidth}
        \centering
        \includegraphics[width=1\textwidth]{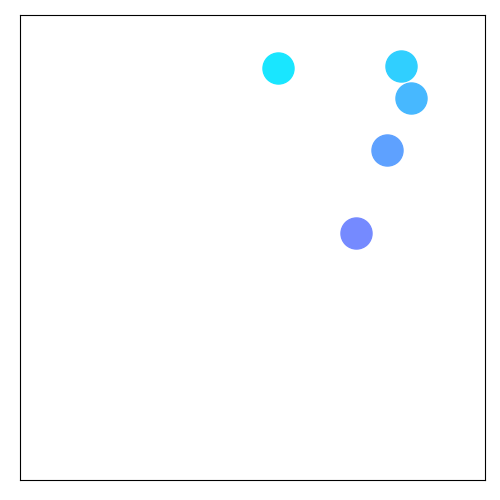}
    \end{subfigure}
        \begin{subfigure}{\csize\textwidth}
        \centering
        \includegraphics[width=1\textwidth]{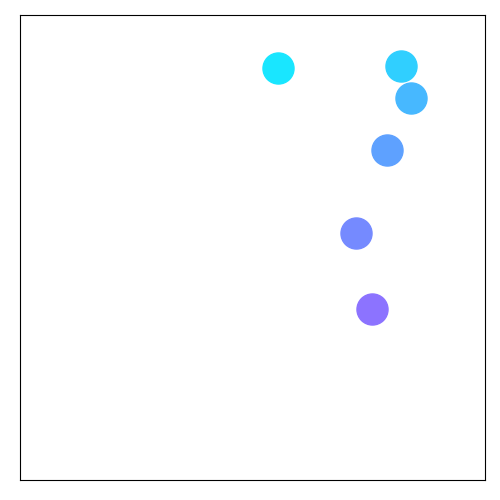}
    \end{subfigure}
        \begin{subfigure}{\csize\textwidth}
        \centering
        \includegraphics[width=1\textwidth]{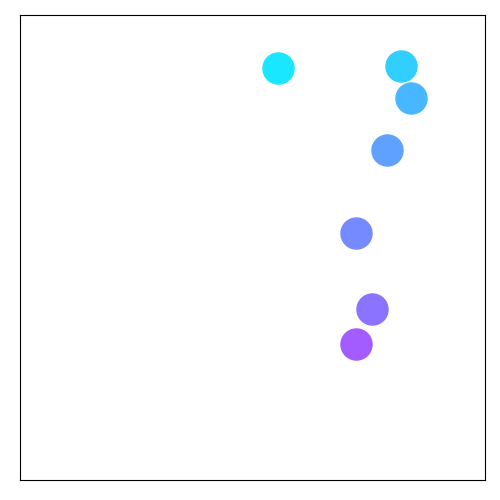}
    \end{subfigure}
        \begin{subfigure}{\csize\textwidth}
        \centering
        \includegraphics[width=1\textwidth]{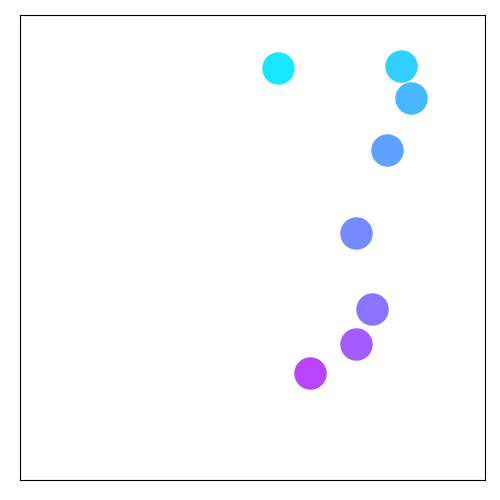}
    \end{subfigure}
        \begin{subfigure}{\csize\textwidth}
        \centering
        \includegraphics[width=1\textwidth]{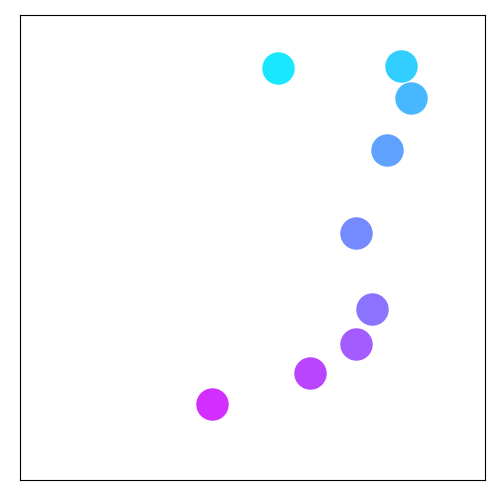}
    \end{subfigure}
        \begin{subfigure}{\csize\textwidth}
        \centering
        \includegraphics[width=1\textwidth]{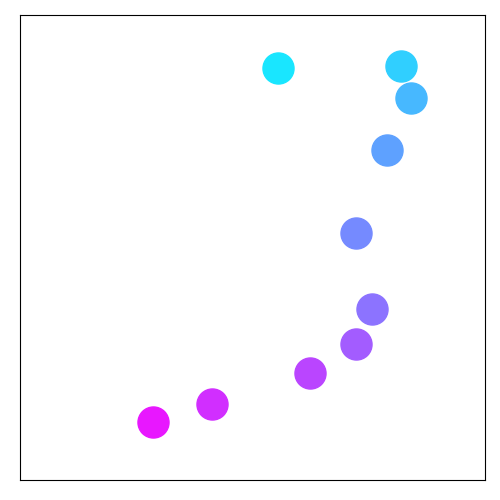}
    \end{subfigure}
        \begin{subfigure}{\csize\textwidth}
        \centering
        \includegraphics[width=1\textwidth]{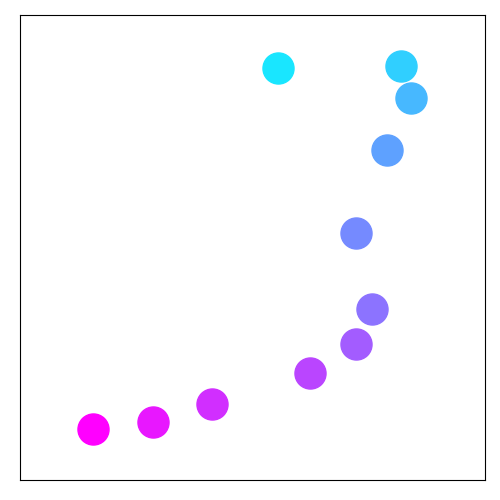}
    \end{subfigure}
    
    \begin{subfigure}{\csize\textwidth}
        \centering
        \textbf{Slot 2:}
    \end{subfigure}
    \begin{subfigure}{\csize\textwidth}
        \centering
        \includegraphics[width=1\textwidth]{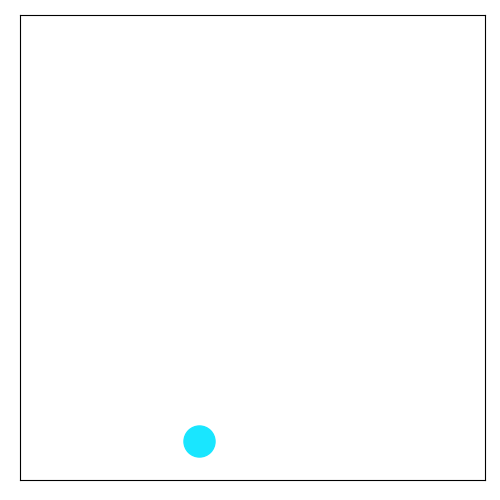}
    \end{subfigure}
        \begin{subfigure}{\csize\textwidth}
        \centering
        \includegraphics[width=1\textwidth]{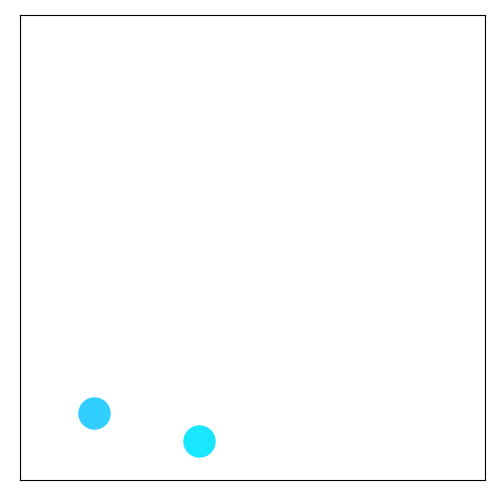}
    \end{subfigure}
        \begin{subfigure}{\csize\textwidth}
        \centering
        \includegraphics[width=1\textwidth]{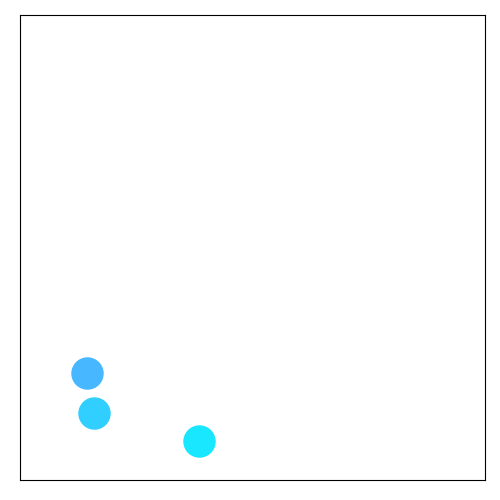}
    \end{subfigure}
        \begin{subfigure}{\csize\textwidth}
        \centering
        \includegraphics[width=1\textwidth]{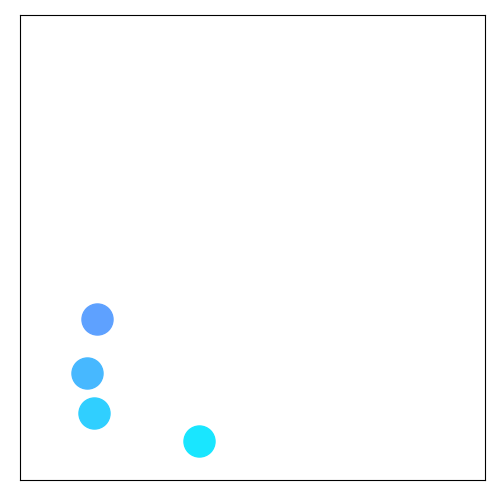}
    \end{subfigure}
        \begin{subfigure}{\csize\textwidth}
        \centering
        \includegraphics[width=1\textwidth]{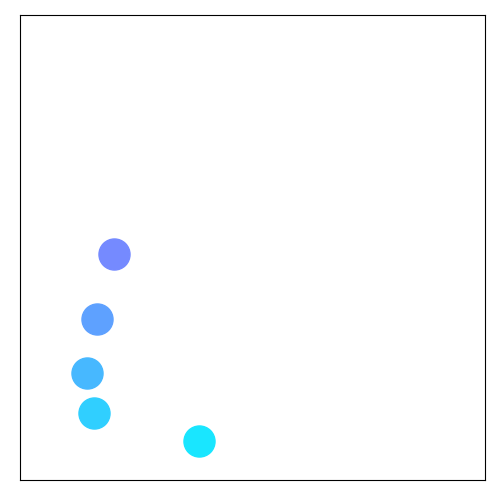}
    \end{subfigure}
        \begin{subfigure}{\csize\textwidth}
        \centering
        \includegraphics[width=1\textwidth]{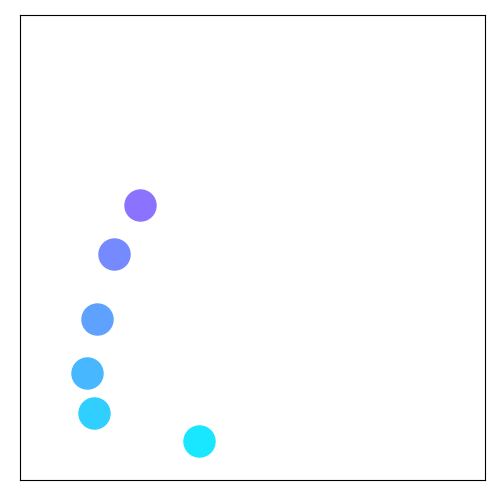}
    \end{subfigure}
        \begin{subfigure}{\csize\textwidth}
        \centering
        \includegraphics[width=1\textwidth]{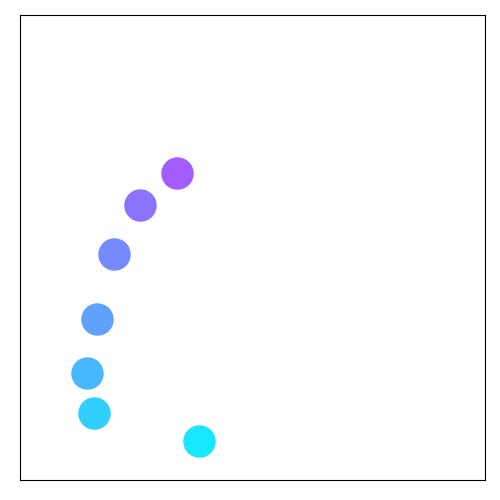}
    \end{subfigure}
        \begin{subfigure}{\csize\textwidth}
        \centering
        \includegraphics[width=1\textwidth]{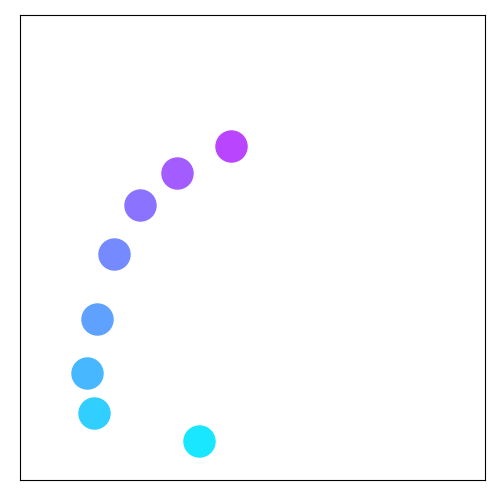}
    \end{subfigure}
        \begin{subfigure}{\csize\textwidth}
        \centering
        \includegraphics[width=1\textwidth]{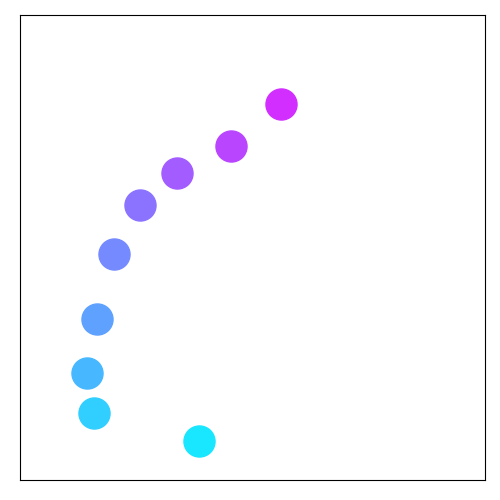}
    \end{subfigure}
        \begin{subfigure}{\csize\textwidth}
        \centering
        \includegraphics[width=1\textwidth]{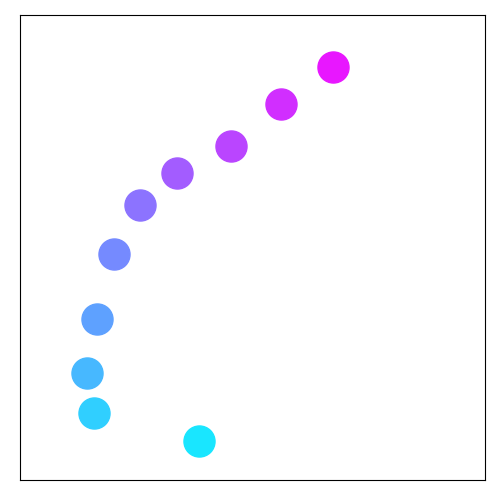}
    \end{subfigure}
        \begin{subfigure}{\csize\textwidth}
        \centering
        \includegraphics[width=1\textwidth]{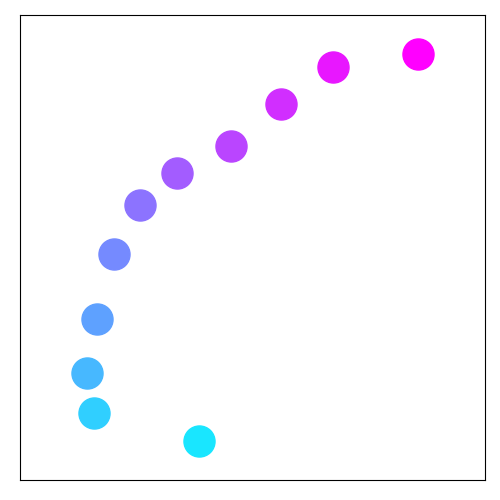}
    \end{subfigure}
    
    \begin{subfigure}{\csize\textwidth}
        \centering
        \textbf{Slot 3:}
    \end{subfigure}
    \begin{subfigure}{\csize\textwidth}
        \centering
        \includegraphics[width=1\textwidth]{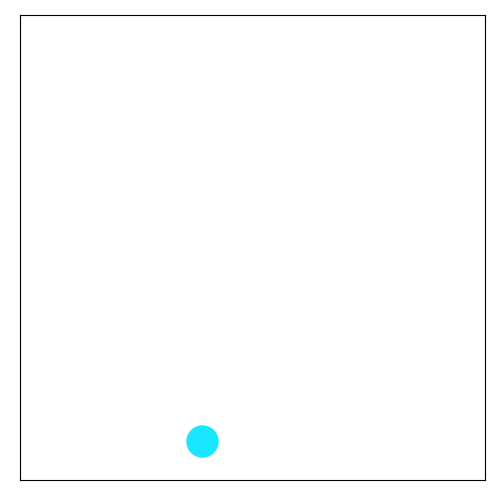}
    \end{subfigure}
        \begin{subfigure}{\csize\textwidth}
        \centering
        \includegraphics[width=1\textwidth]{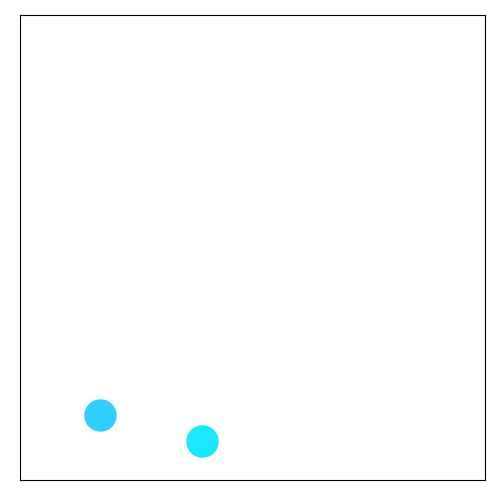}
    \end{subfigure}
        \begin{subfigure}{\csize\textwidth}
        \centering
        \includegraphics[width=1\textwidth]{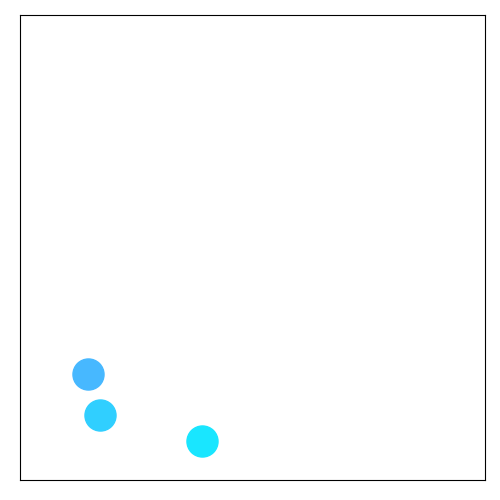}
    \end{subfigure}
        \begin{subfigure}{\csize\textwidth}
        \centering
        \includegraphics[width=1\textwidth]{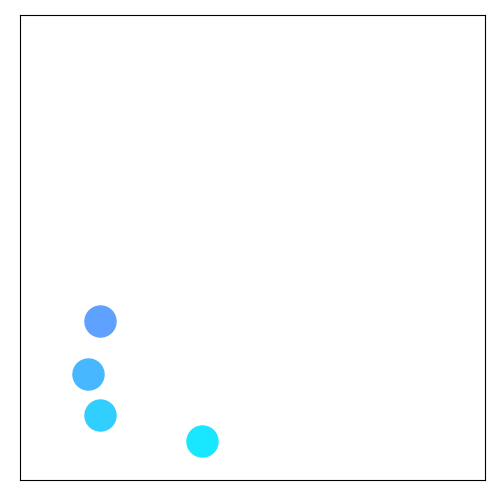}
    \end{subfigure}
        \begin{subfigure}{\csize\textwidth}
        \centering
        \includegraphics[width=1\textwidth]{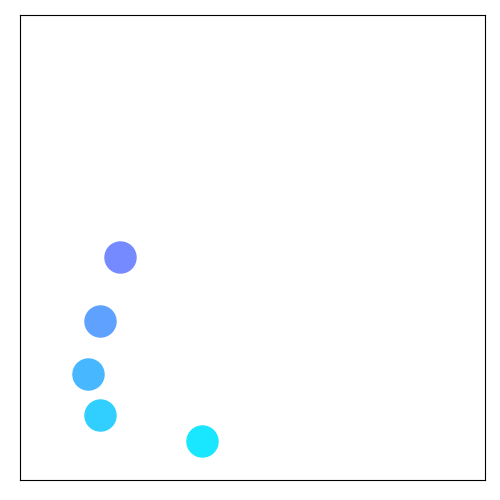}
    \end{subfigure}
        \begin{subfigure}{\csize\textwidth}
        \centering
        \includegraphics[width=1\textwidth]{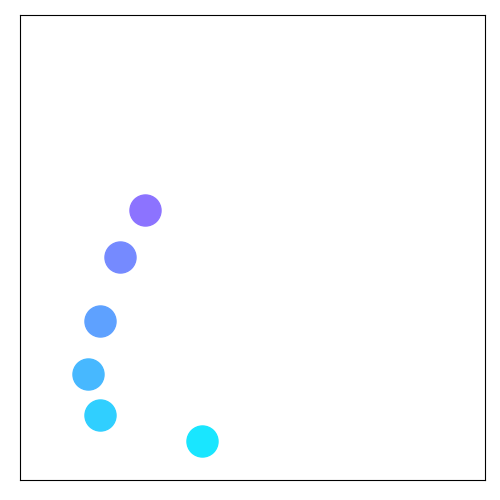}
    \end{subfigure}
        \begin{subfigure}{\csize\textwidth}
        \centering
        \includegraphics[width=1\textwidth]{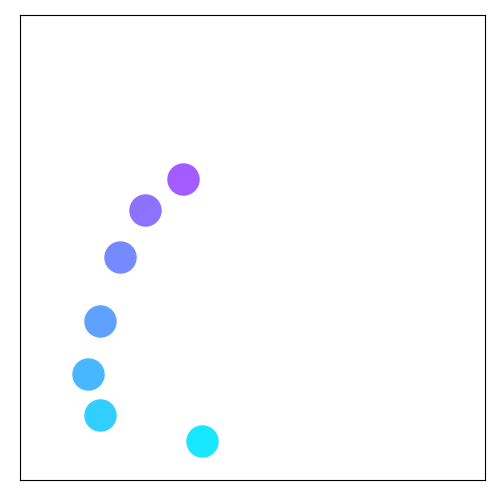}
    \end{subfigure}
        \begin{subfigure}{\csize\textwidth}
        \centering
        \includegraphics[width=1\textwidth]{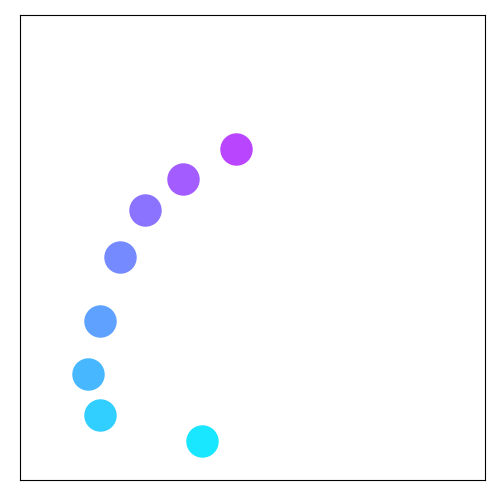}
    \end{subfigure}
        \begin{subfigure}{\csize\textwidth}
        \centering
        \includegraphics[width=1\textwidth]{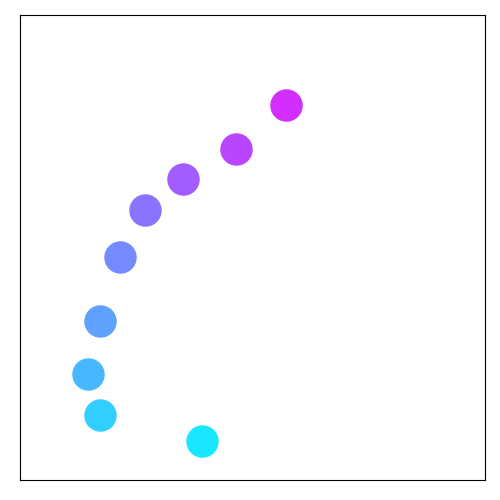}
    \end{subfigure}
        \begin{subfigure}{\csize\textwidth}
        \centering
        \includegraphics[width=1\textwidth]{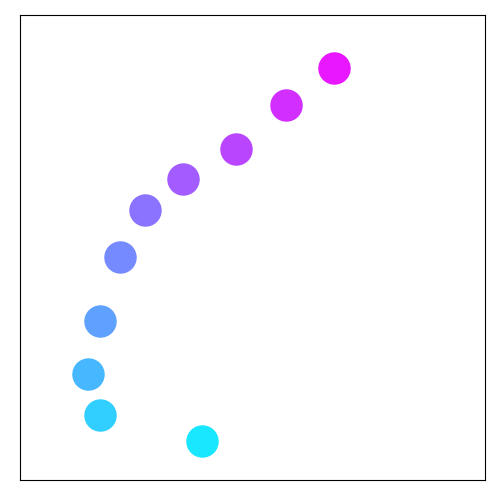}
    \end{subfigure}
        \begin{subfigure}{\csize\textwidth}
        \centering
        \includegraphics[width=1\textwidth]{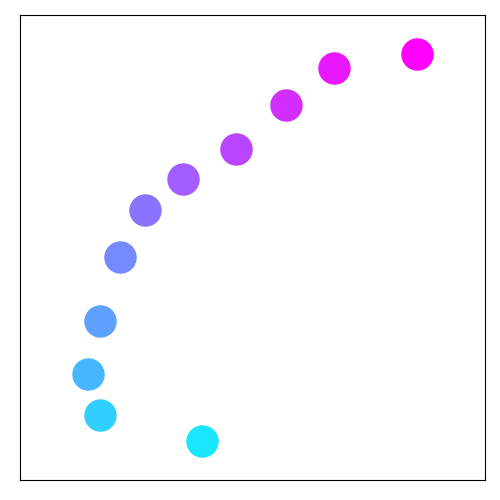}
    \end{subfigure}
    
    \caption{Visualization of encoded states for a single episode of Atari Pong. Each state is encoded into a latent state with three slots, and we color encodings from blue to purple as the time progresses.}
    \label{fig:pong_correlation}
\end{figure}

\subsection{Evaluation Metrics}
\label{ap:metrics}

\textbf{C-SWM Hits@1 and Mean Reciprocal Rank:} We follow the evaluation setup from \citet{kipf20contrastive}. An evaluation dataset of episodes of length 10 is collected. The agent's encoder is used to encode all states in the evaluation dataset into latent states. Then the agent predict the next latent state $t=1$, $t=5$ or $t=10$ steps into the future given a sequences of 1, 5 or 10 actions respectively. The predicted latent state is then compared to the actual encoded state at timestep $t$ (the positive example) and to all states at timestep $t$ from other episodes (the negative example). The Hits@1 metric simply counts the fraction of times the predicted latent state was closer to the positive example than to all other negative examples. The Mean Reciprocal Rank (MRR) metric computes
\begin{align}
    \frac{1}{|D|} \sum_{n=1}^{|D|} \frac{1}{\text{rank}_n}, 
\end{align}
where $D$ is the evaluation dataset and $\text{rank}_n$ is the index of the positive examples in the sorted list of distances between the predicted latent state and the set of the positive and all negative examples.

The original evaluation code does not always handle duplicate states in different episodes well. It is assumed that two identical images will be encoded into the same latent state (i.e. all float32 values will be exactly equal), but this does not happen in practice even when pytorch is set to a deterministic mode. Duplicate states are infrequent in \texttt{2D Shapes} and \texttt{3D Cubes}, but fairly common in Atari. In the Atari experiments, we check for duplicate states and remove negative states that are identical to a particular positive state.

\textbf{Slot Correlation:} we compute the absolute value of the Pearson correlation coefficient between every pair of object slots. This value is averaged over all pairs of object slots and over all states in the evaluation set. The resulting value is bounded between zero and one, where zero means no correlation between slots and one means that all slots capture the same information.

\textbf{FWM Block Position RMSE and Action Sequence Hits@1:} We use the same metrics as \citet{biza2022factored}. To the block position prediction, we train a Multi-Layer Perceptron to predict the spatial $(x, y, z)$ position of each block from the latent space of FWMs. Only the MLP is trained, all other weights are frozen. Then, we compute the root mean squared error between the actual and predicted block positions. The dataset used in this evaluation is the dataset of expert demonstrations for the four zero-shot transfer tasks.

In action sequence ranking, we generate one correct and ten incorrect trajectories for each zero-shot transfer task. We also provide the model with an example of a goal state. The model encodes the goal state and predicts the resulting latent state of all eleven action sequences. If the predicted result for the correct trajectory is the closest to the encoded goal state, the prediction of the model is considered correct. Hits@1 once again measures the fraction of correct predictions.

\begin{flushleft}
\begin{table}
    \centering
    \begin{tabular}{lc@{\hskip3pt}c@{\hskip3pt}c@{\hskip3pt}c@{\hskip3pt}c@{\hskip3pt}c@{\hskip3pt}c@{\hskip3pt}c@{\hskip3pt}c@{\hskip3pt}c@{\hskip3pt}c}
        \toprule
        \textbf{Task} & \multicolumn{11}{c}{\textbf{State sequences and attention weights}} \\
        & Init & Pick & Place & Pick & Place & Pick & Place & Pick & Place & Pick & Place \\
        \midrule
        Wall &
        \includegraphics[width=\csize\textwidth]{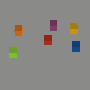} & 
        \includegraphics[width=\csize\textwidth]{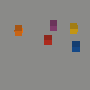} & \includegraphics[width=\csize\textwidth]{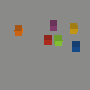} & \includegraphics[width=\csize\textwidth]{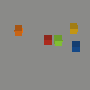} &
        \includegraphics[width=\csize\textwidth]{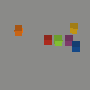} &
        \includegraphics[width=\csize\textwidth]{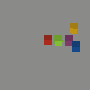} &
        \includegraphics[width=\csize\textwidth]{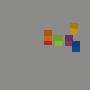} & 
        \includegraphics[width=\csize\textwidth]{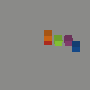} & 
        \includegraphics[width=\csize\textwidth]{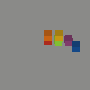} & 
        \includegraphics[width=\csize\textwidth]{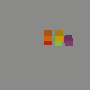} & 
        \includegraphics[width=\csize\textwidth]{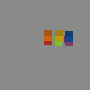} \\
         & &
        \includegraphics[width=\csize\textwidth]{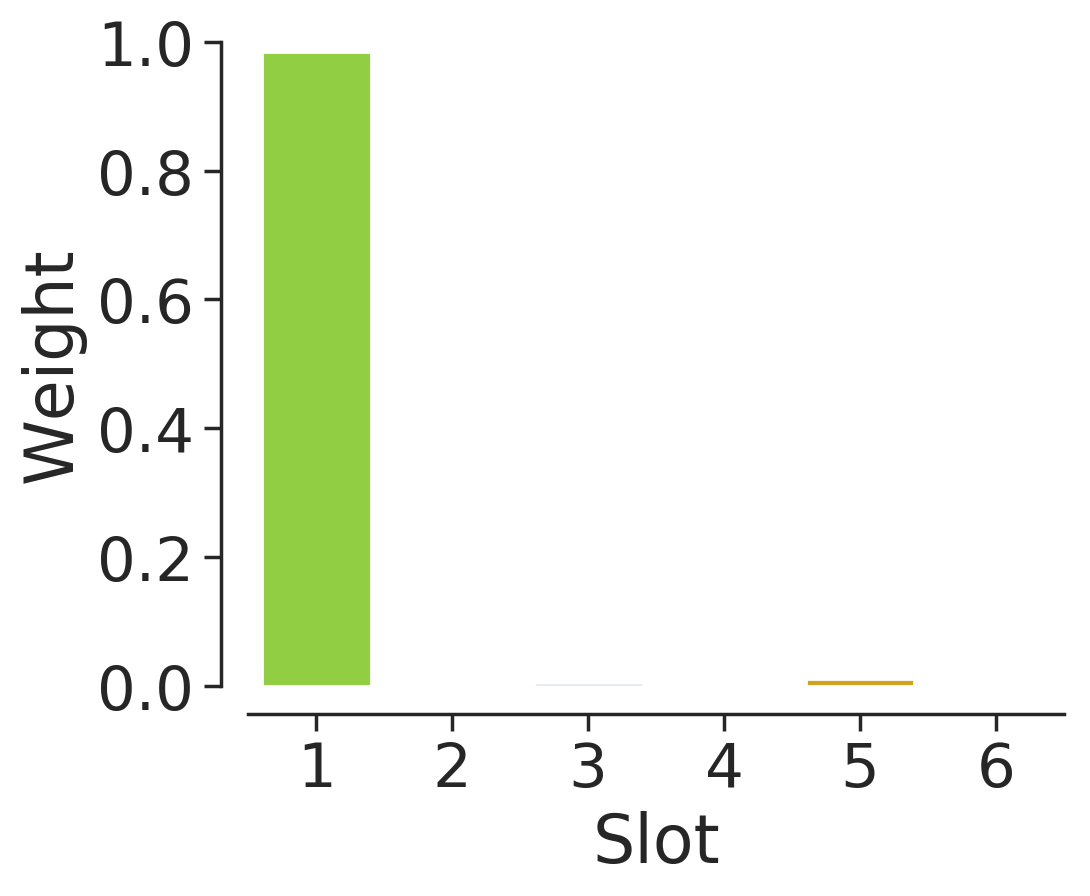} & \includegraphics[width=\csize\textwidth]{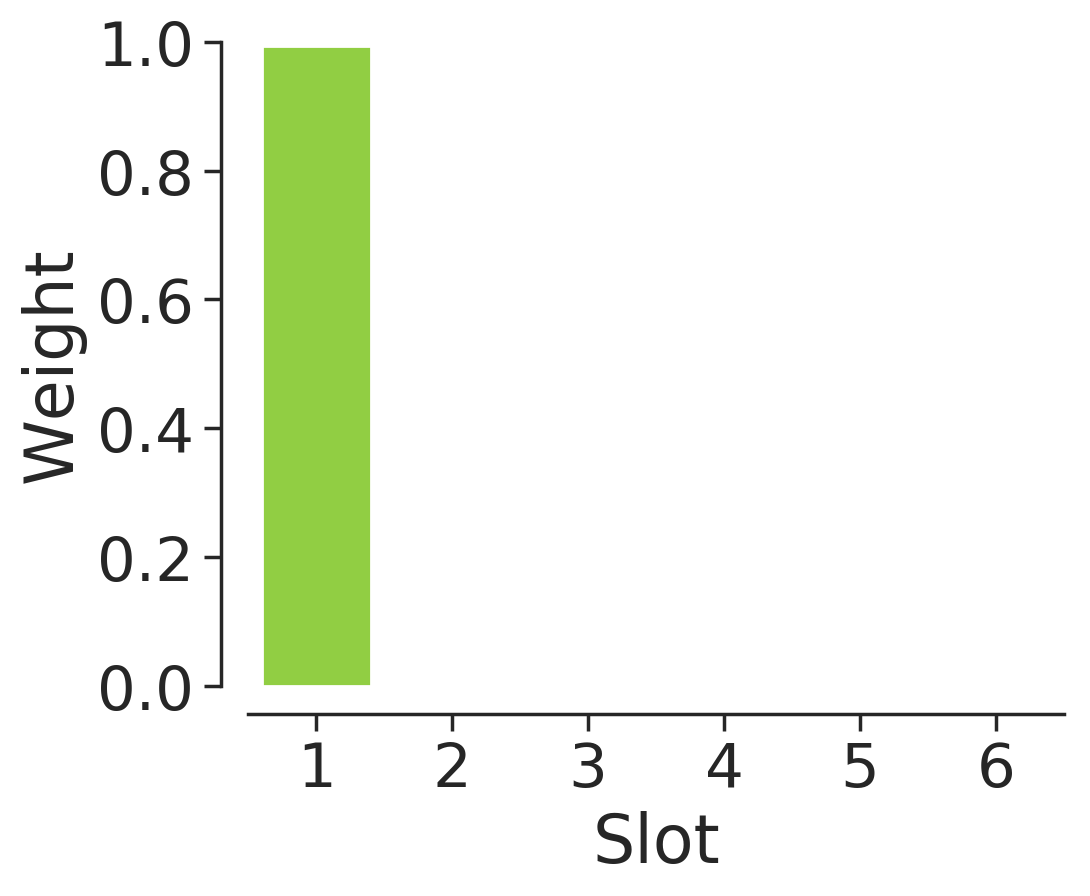} & \includegraphics[width=\csize\textwidth]{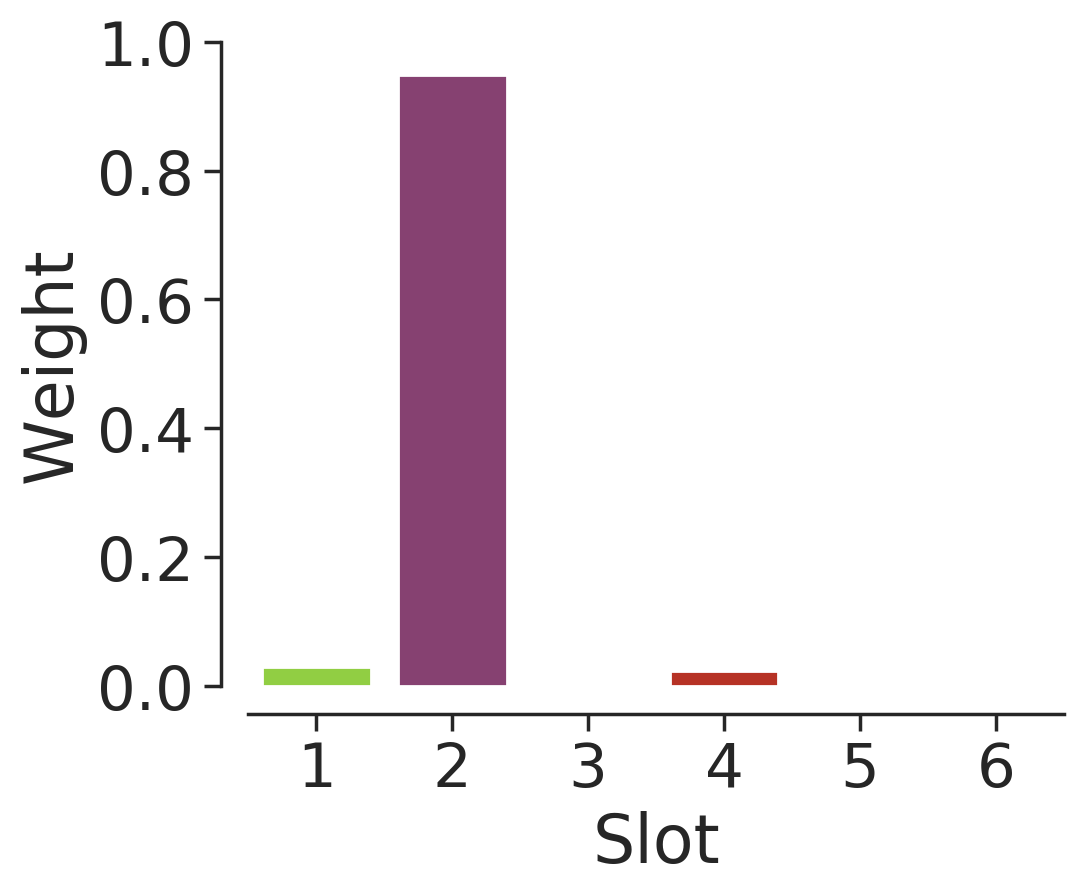} &
        \includegraphics[width=\csize\textwidth]{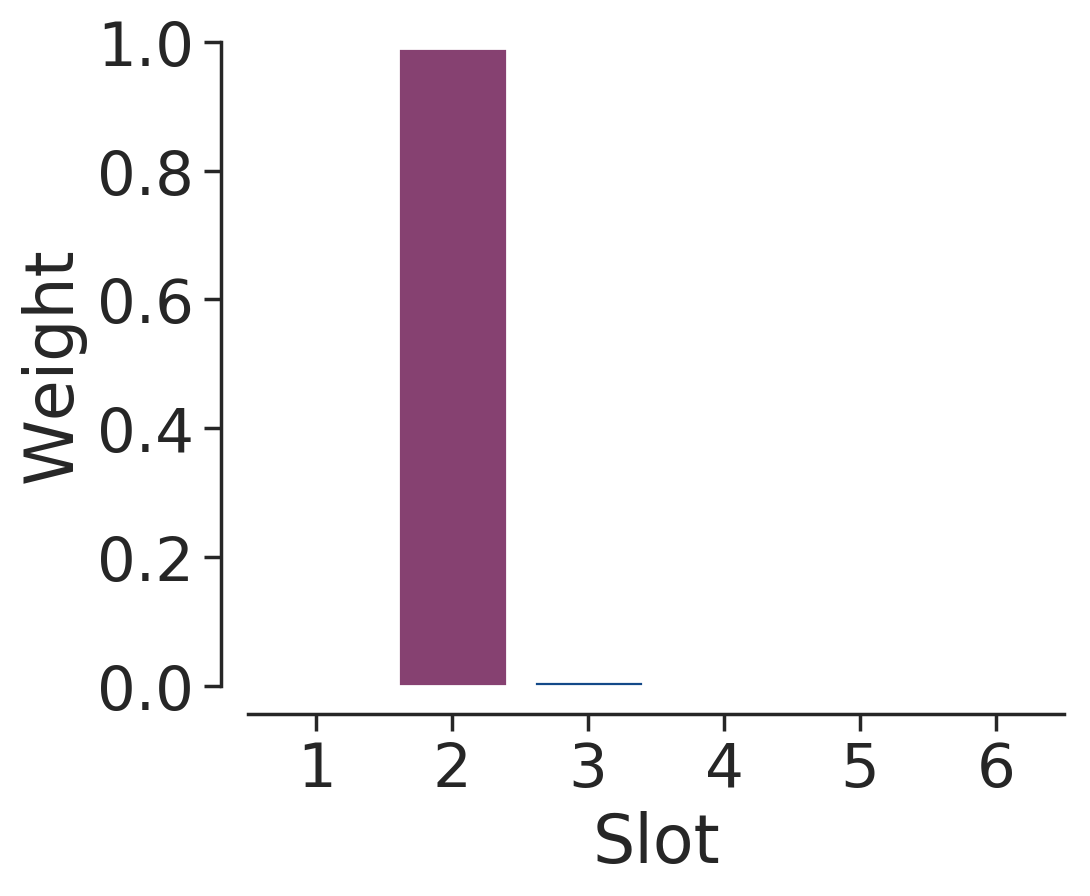} &
        \includegraphics[width=\csize\textwidth]{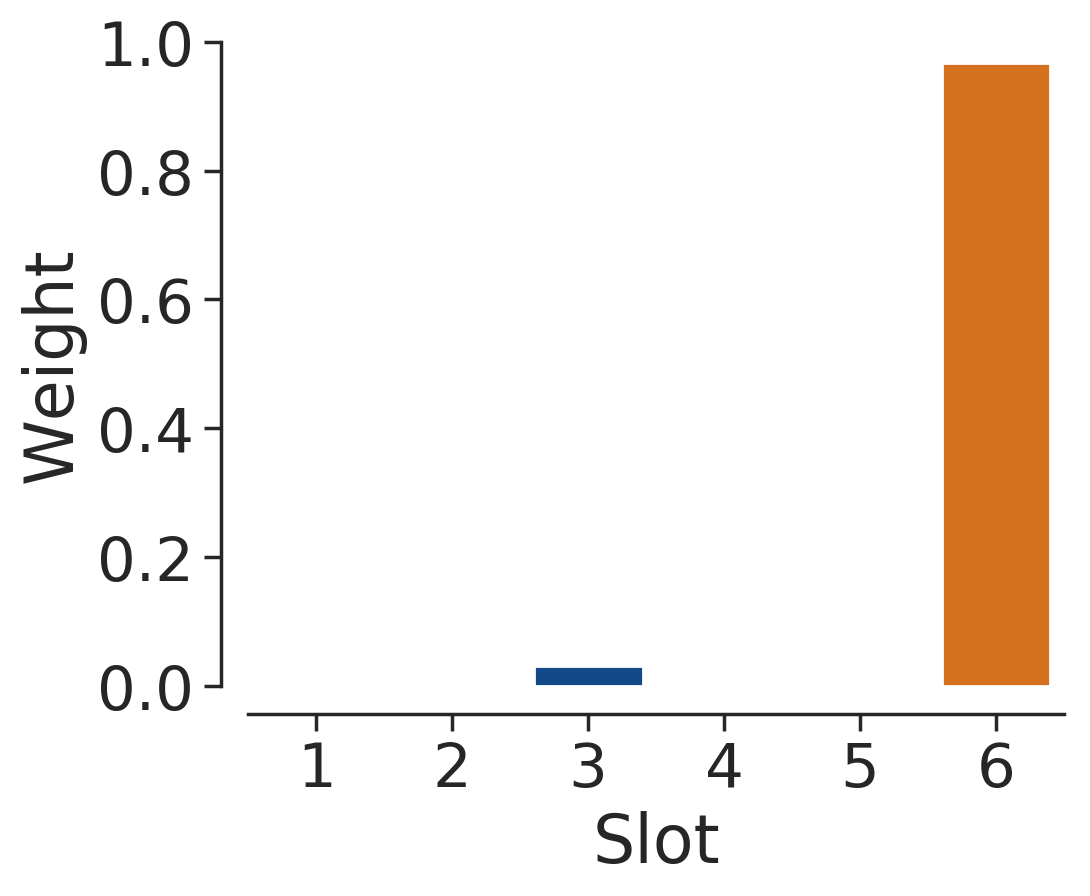} &
        \includegraphics[width=\csize\textwidth]{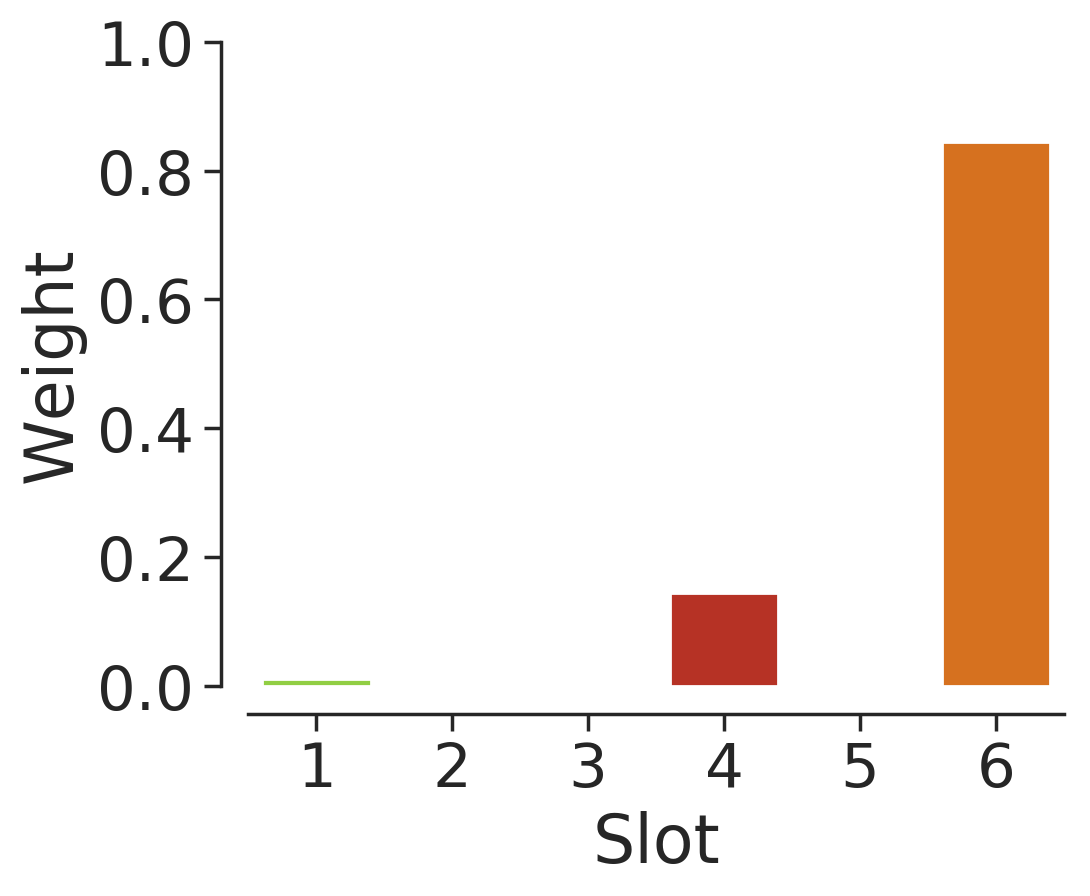} & 
        \includegraphics[width=\csize\textwidth]{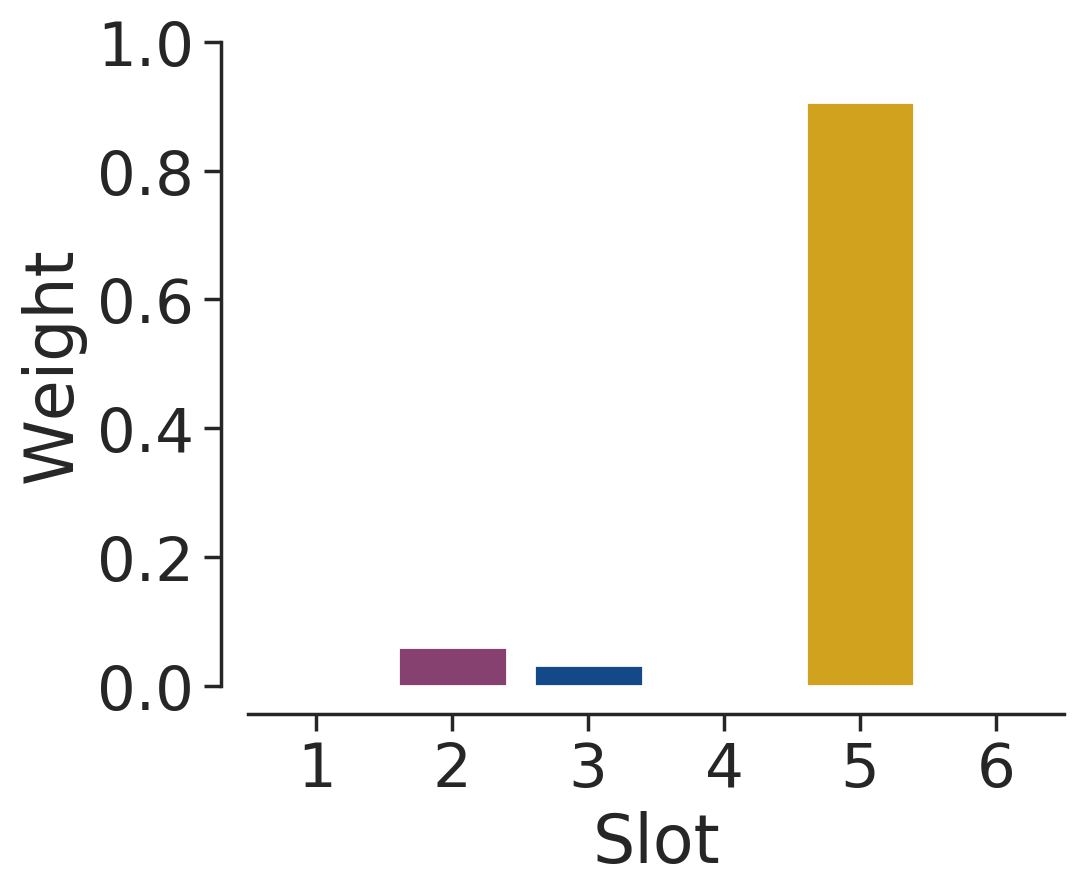} & 
        \includegraphics[width=\csize\textwidth]{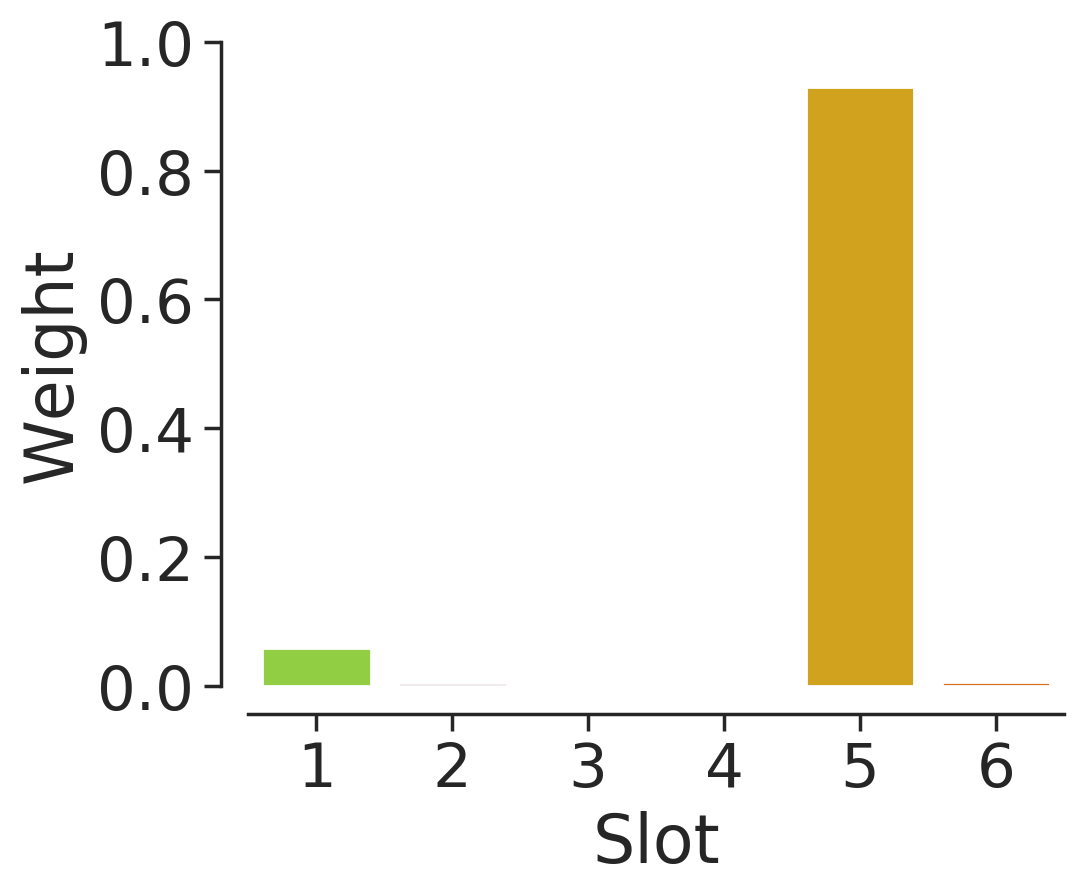} &
        \includegraphics[width=\csize\textwidth]{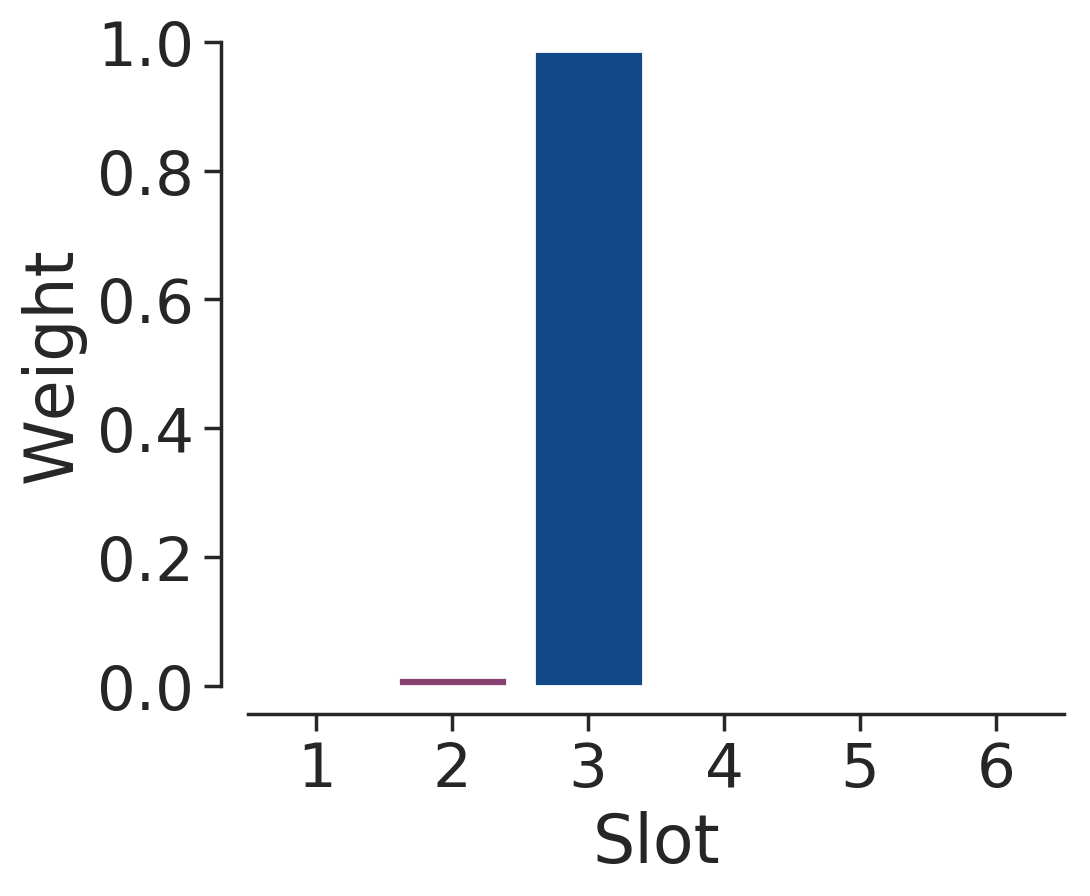} & 
        \includegraphics[width=\csize\textwidth]{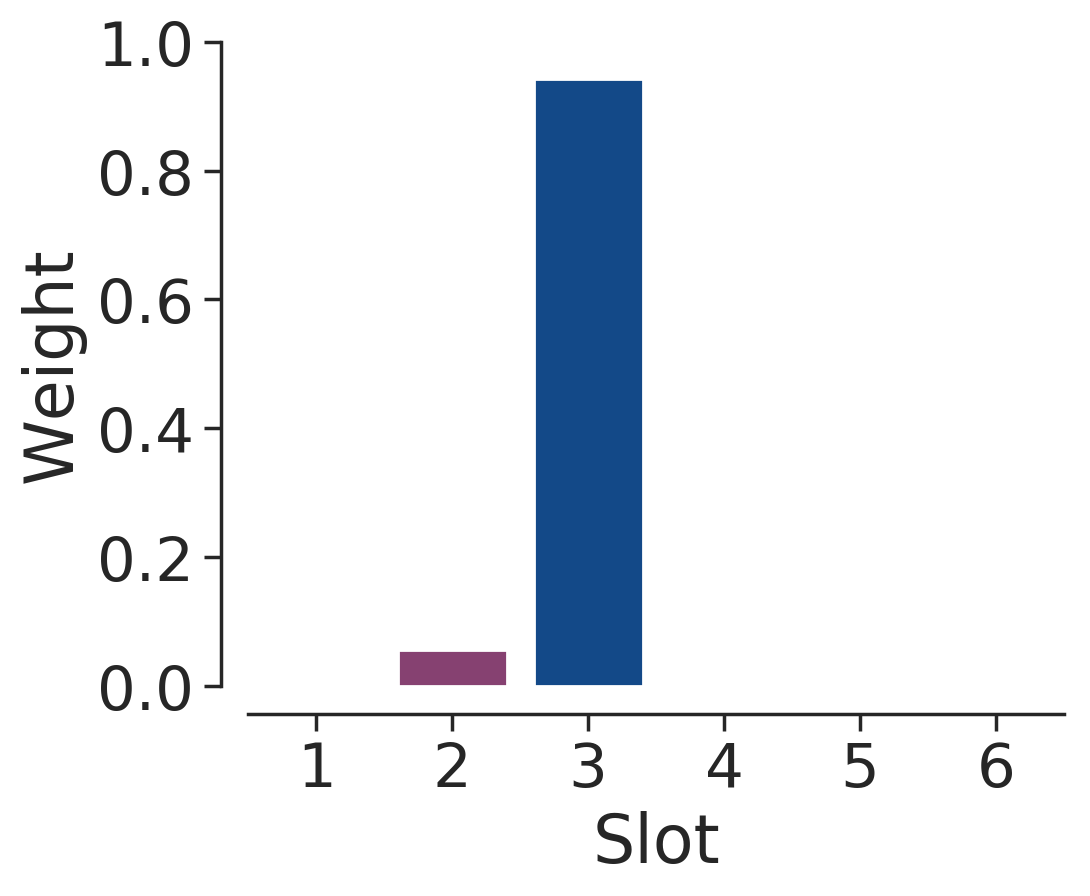} \\
        \midrule
        Stairs &
        \includegraphics[width=\csize\textwidth]{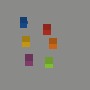} & 
        \includegraphics[width=\csize\textwidth]{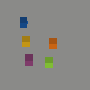} & \includegraphics[width=\csize\textwidth]{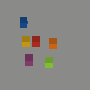} & \includegraphics[width=\csize\textwidth]{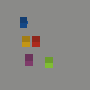} &
        \includegraphics[width=\csize\textwidth]{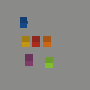} &
        \includegraphics[width=\csize\textwidth]{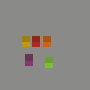} &
        \includegraphics[width=\csize\textwidth]{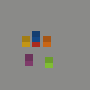} & 
        \includegraphics[width=\csize\textwidth]{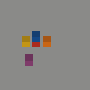} & 
        \includegraphics[width=\csize\textwidth]{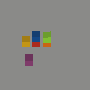} & 
        \includegraphics[width=\csize\textwidth]{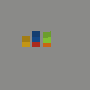} & 
        \includegraphics[width=\csize\textwidth]{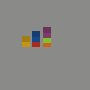} \\
         & &
        \includegraphics[width=\csize\textwidth]{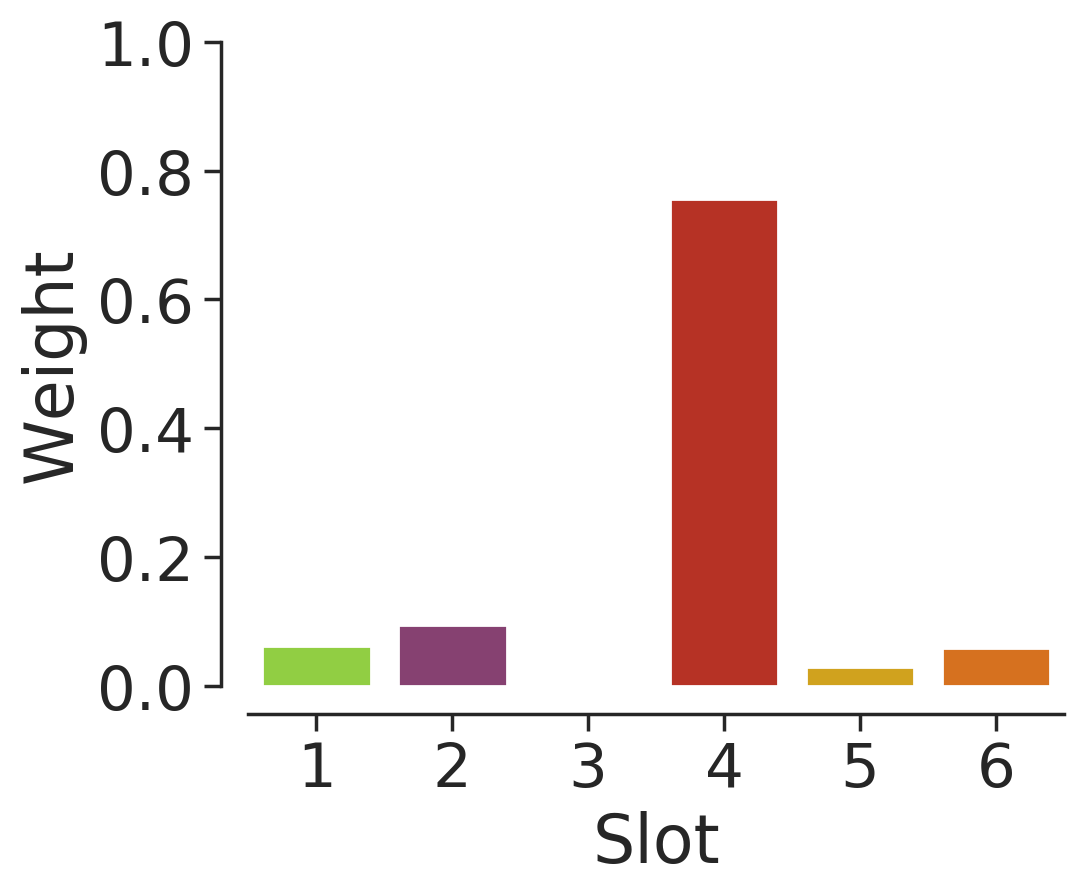} & \includegraphics[width=\csize\textwidth]{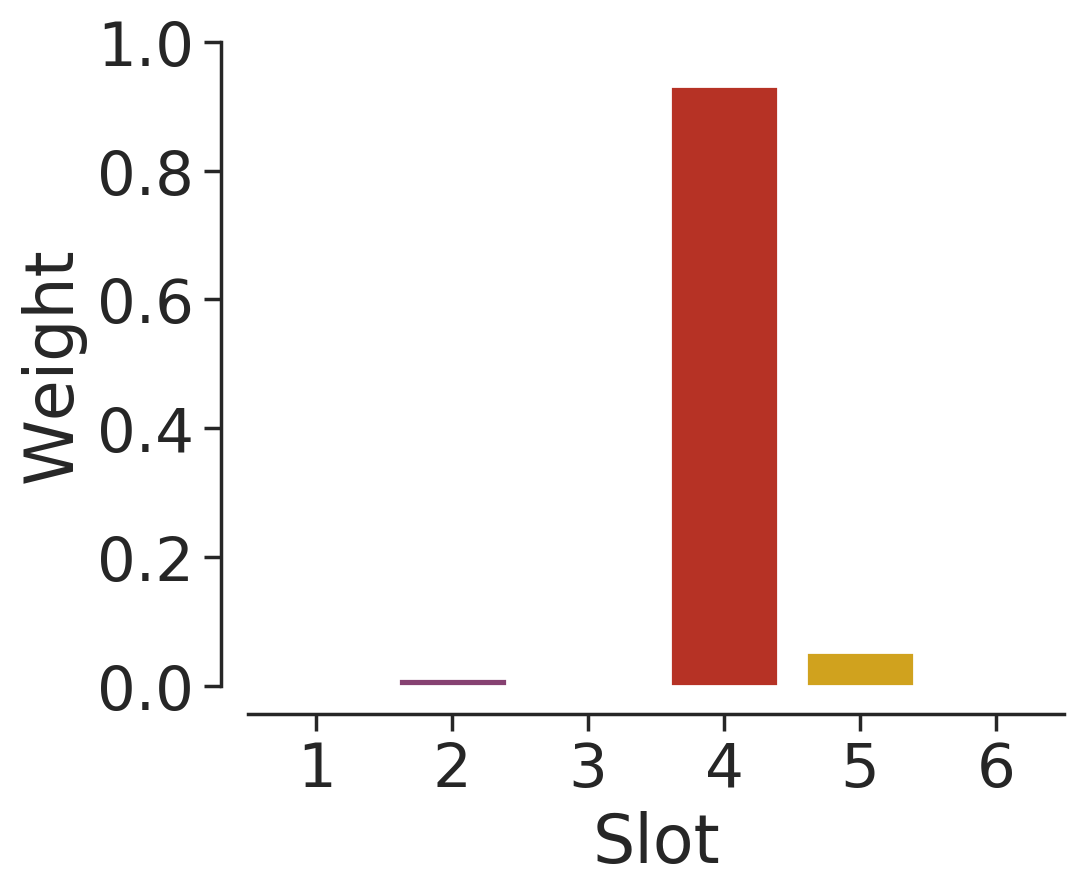} & \includegraphics[width=\csize\textwidth]{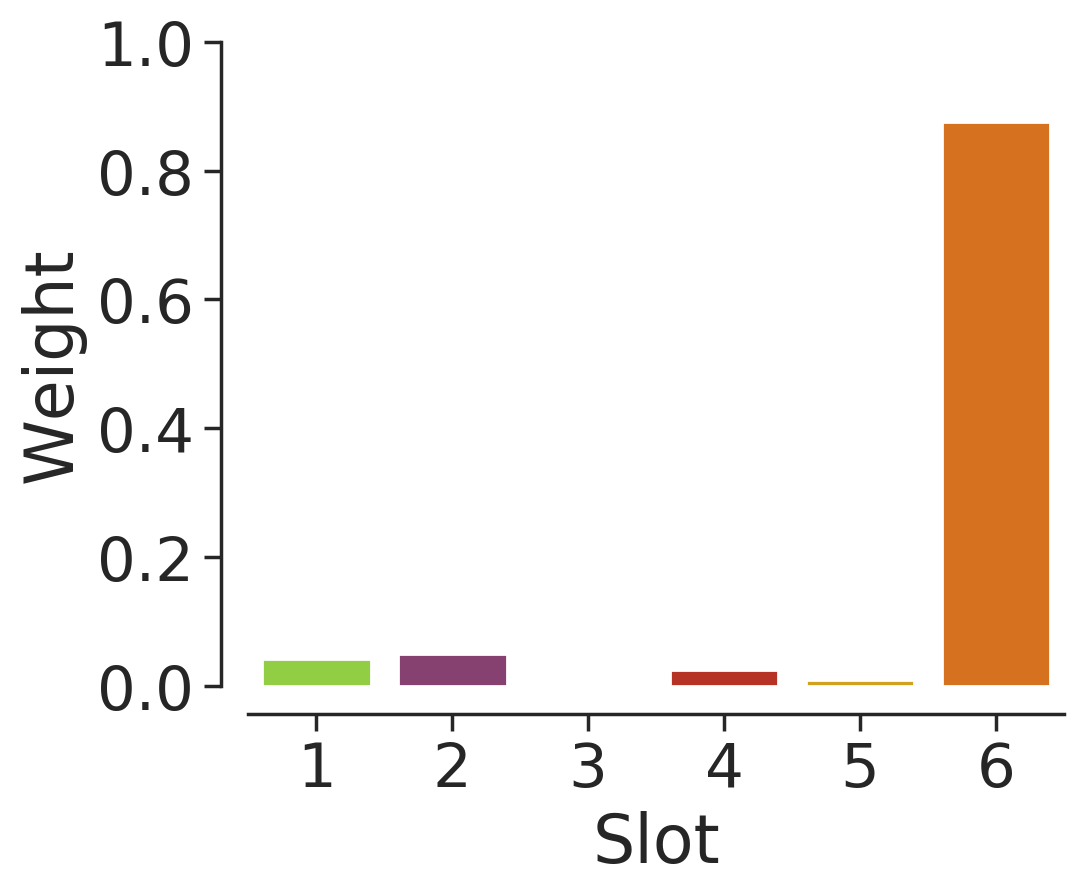} &
        \includegraphics[width=\csize\textwidth]{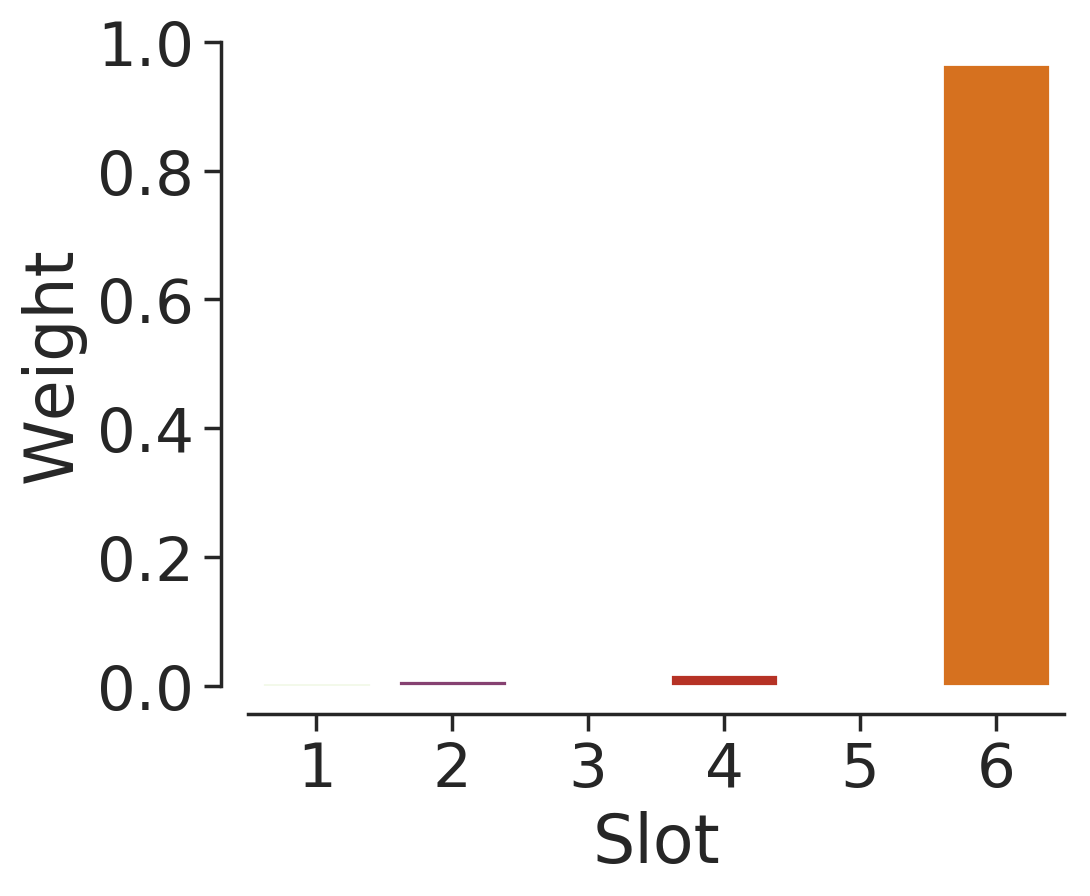} &
        \includegraphics[width=\csize\textwidth]{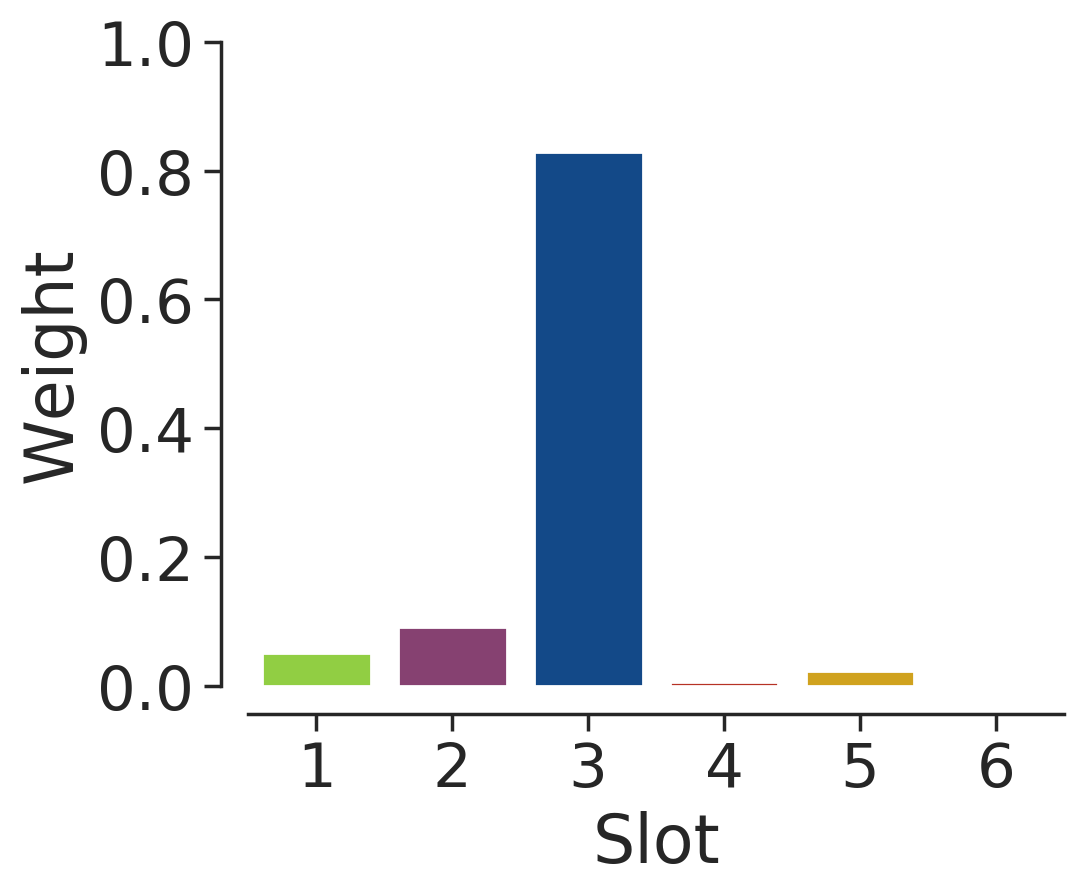} &
        \includegraphics[width=\csize\textwidth]{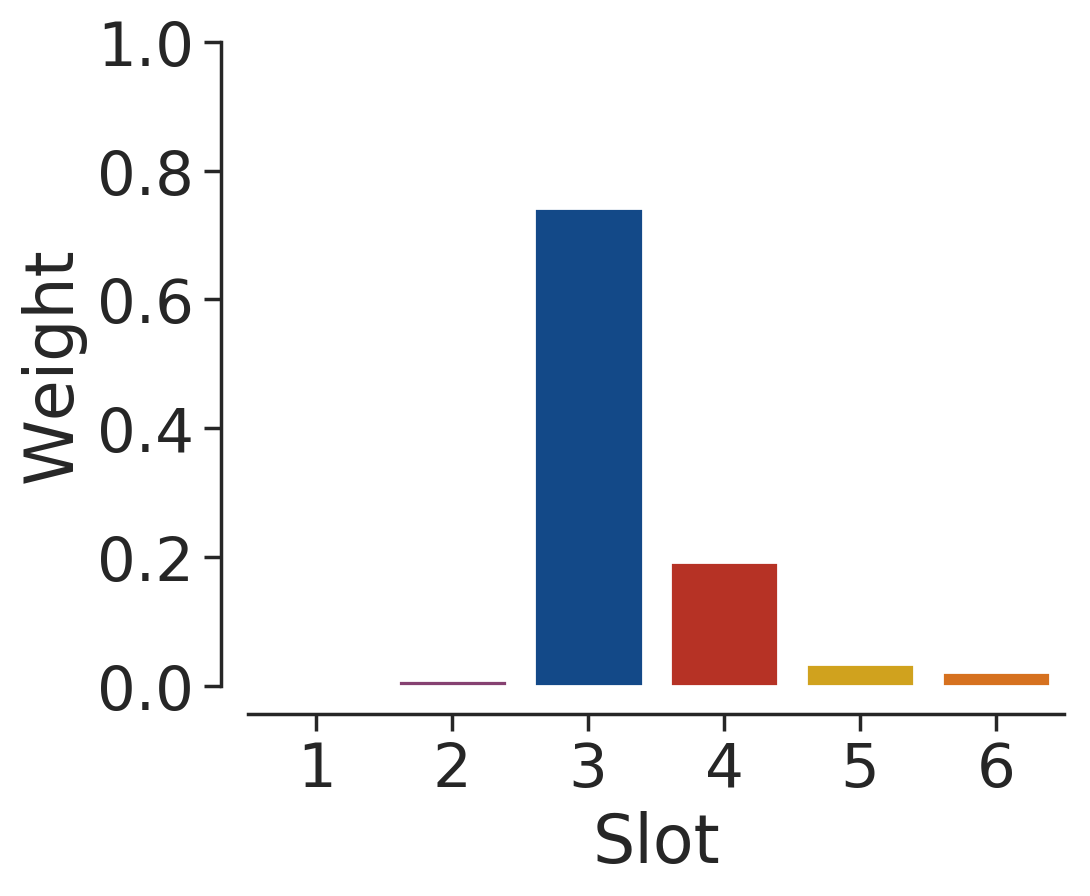} & 
        \includegraphics[width=\csize\textwidth]{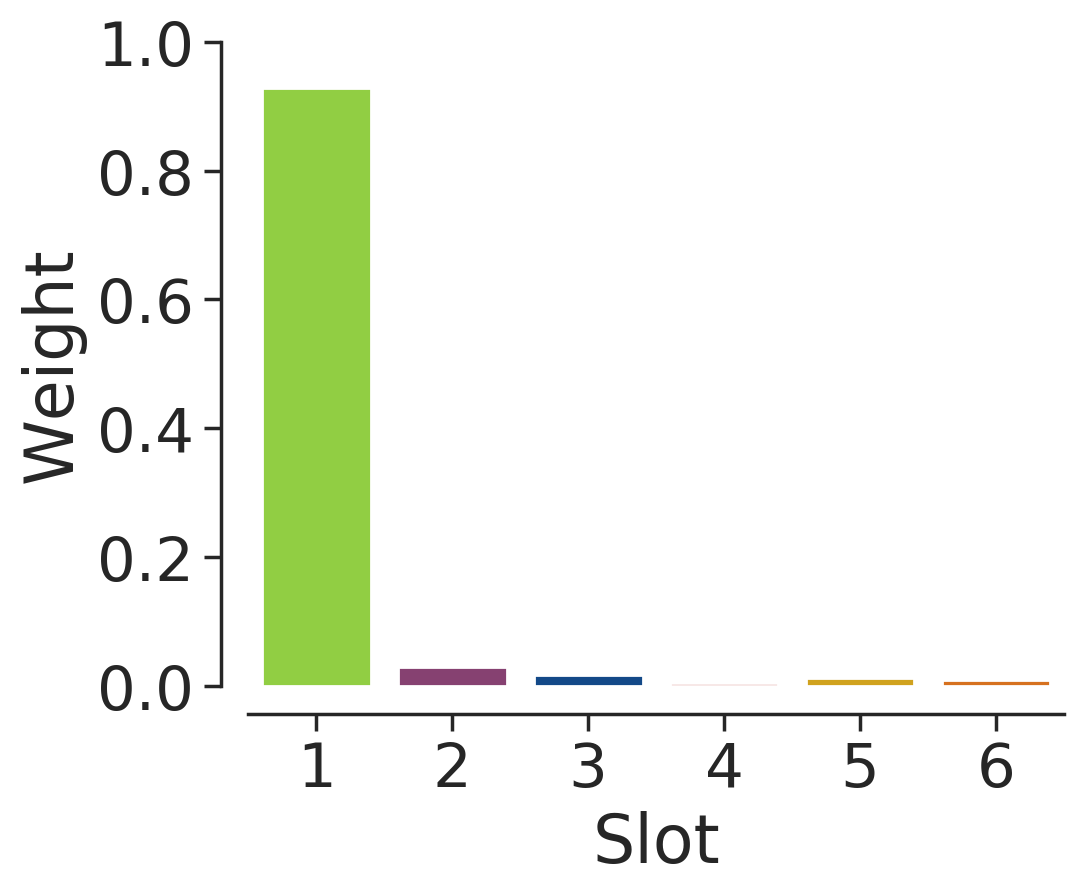} & 
        \includegraphics[width=\csize\textwidth]{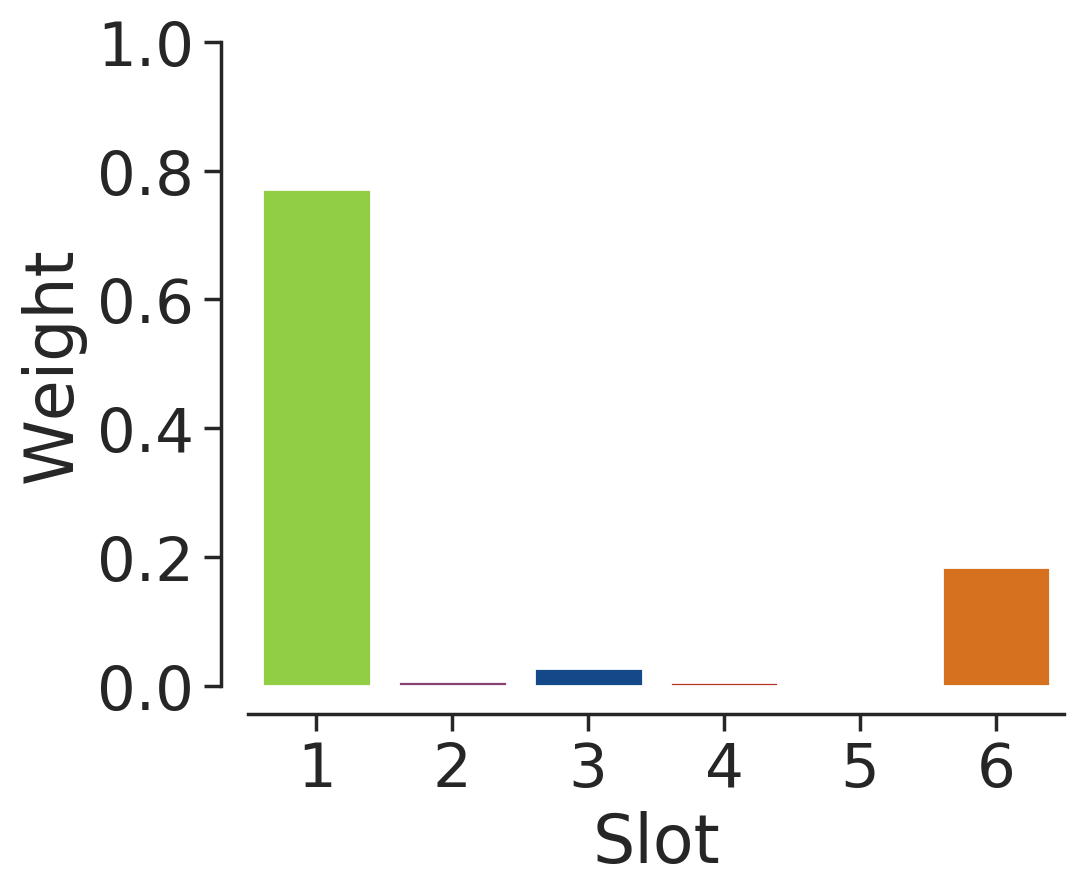} &
        \includegraphics[width=\csize\textwidth]{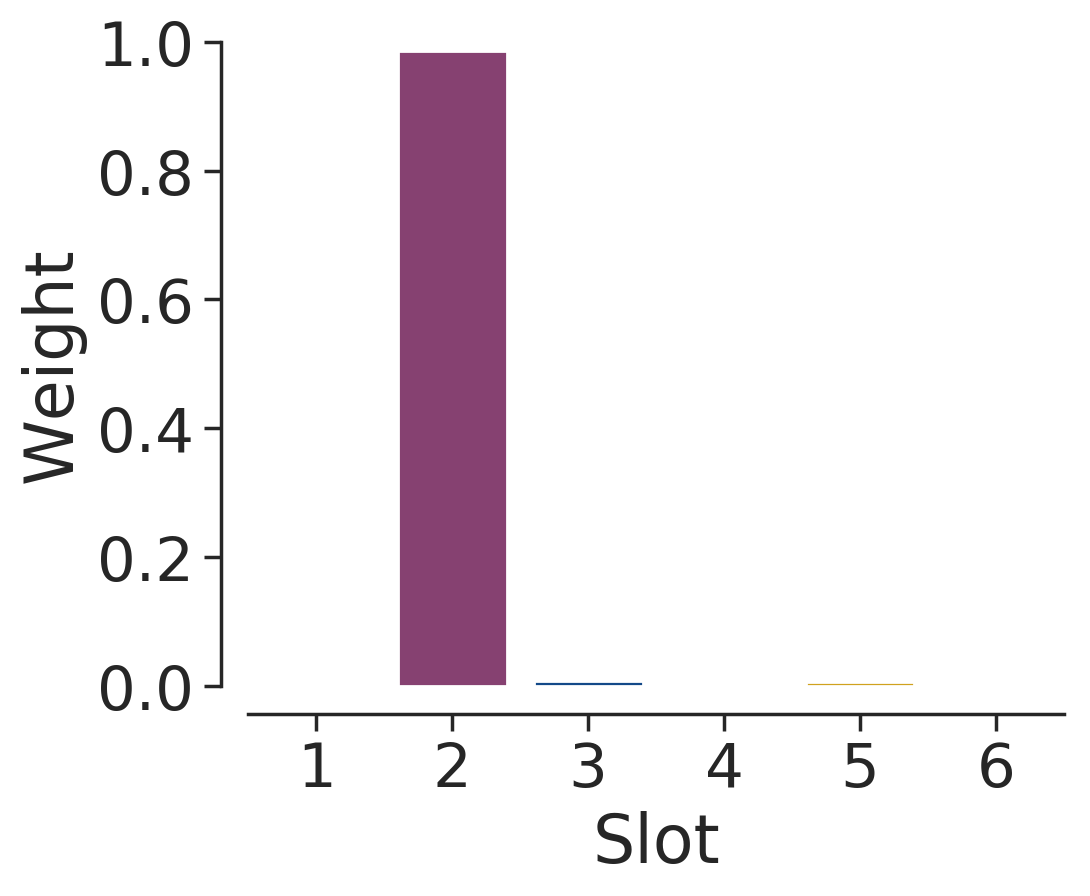} & 
        \includegraphics[width=\csize\textwidth]{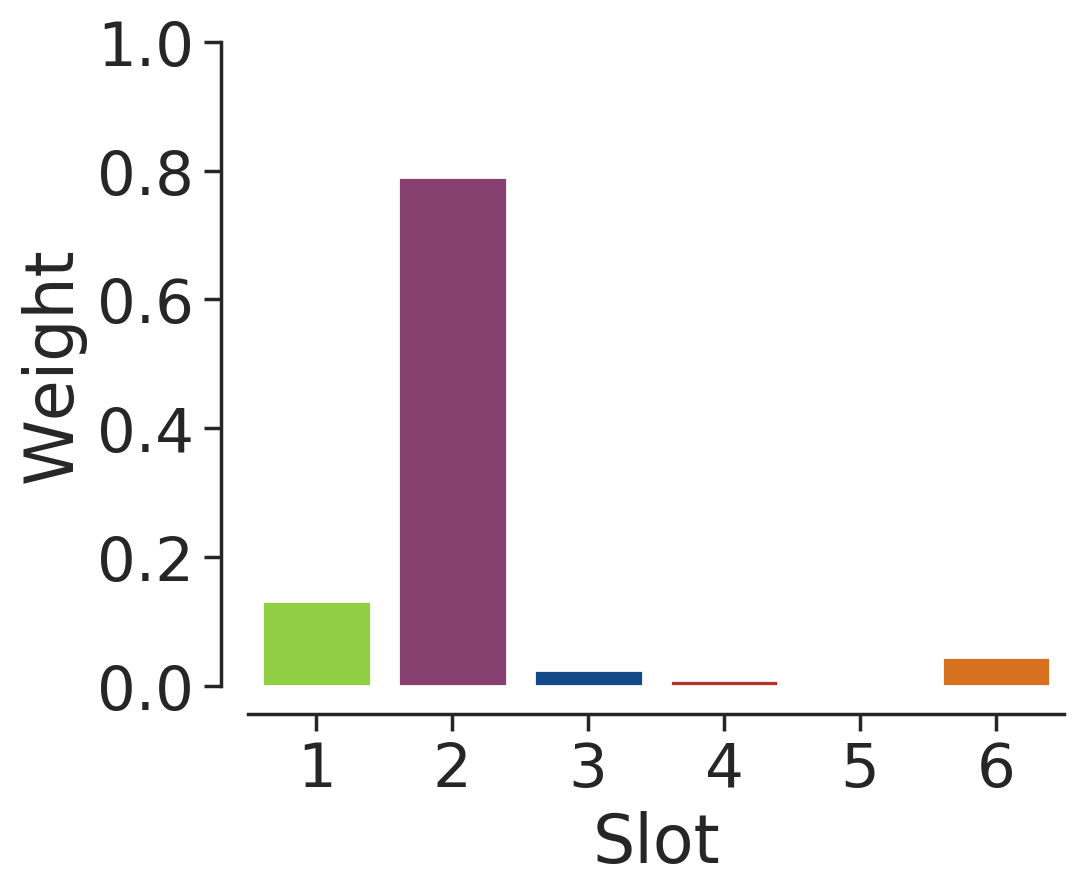} \\
        \midrule
        2 towers &
        \includegraphics[width=\csize\textwidth]{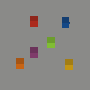} & 
        \includegraphics[width=\csize\textwidth]{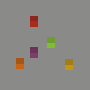} & \includegraphics[width=\csize\textwidth]{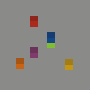} & \includegraphics[width=\csize\textwidth]{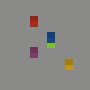} &
        \includegraphics[width=\csize\textwidth]{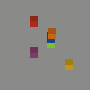} &
        \includegraphics[width=\csize\textwidth]{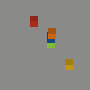} &
        \includegraphics[width=\csize\textwidth]{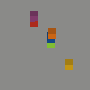} & 
        \includegraphics[width=\csize\textwidth]{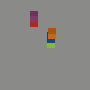} & 
        \includegraphics[width=\csize\textwidth]{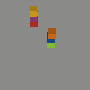} & 
         & \\
         & &
        \includegraphics[width=\csize\textwidth]{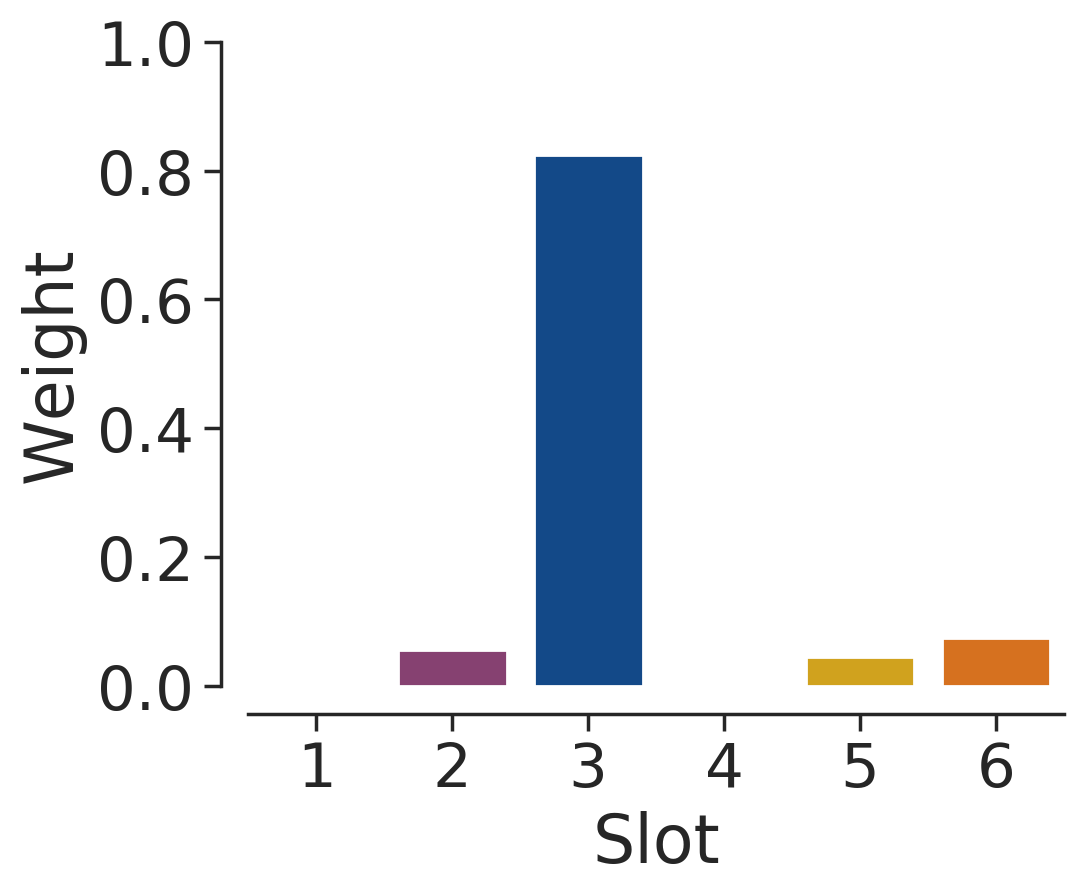} & \includegraphics[width=\csize\textwidth]{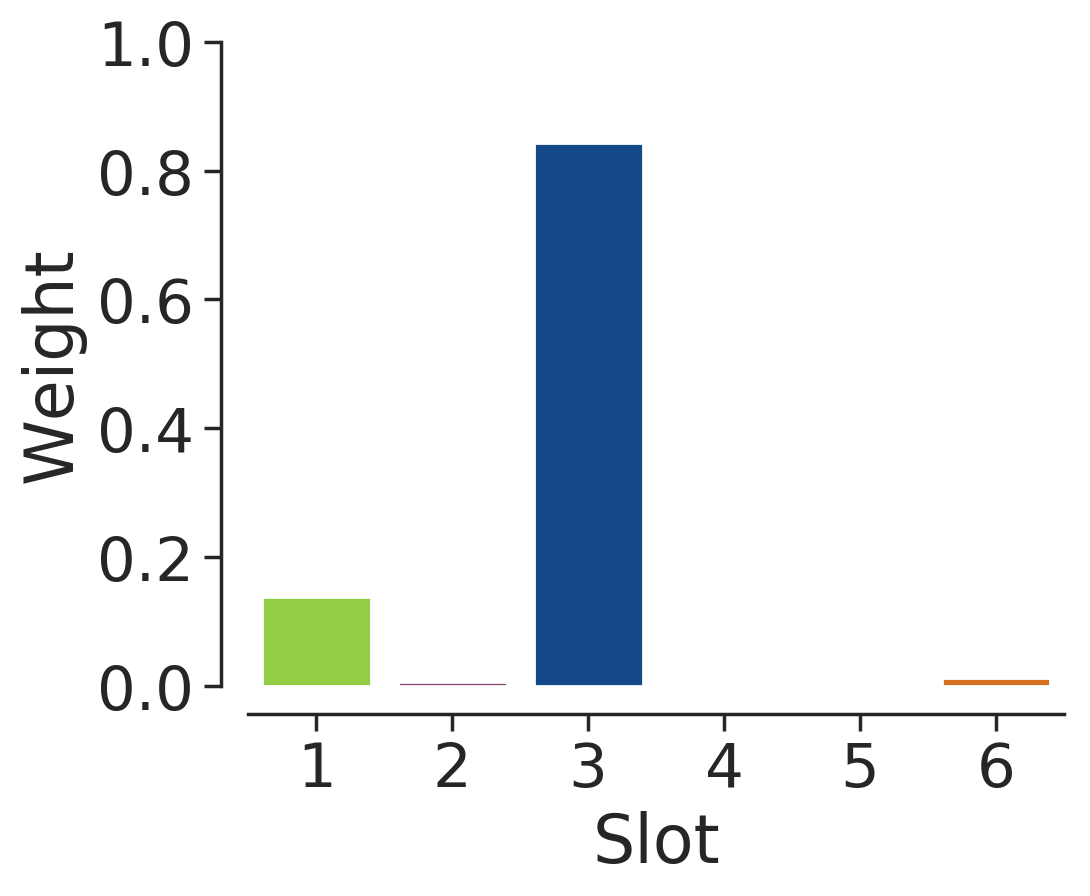} & \includegraphics[width=\csize\textwidth]{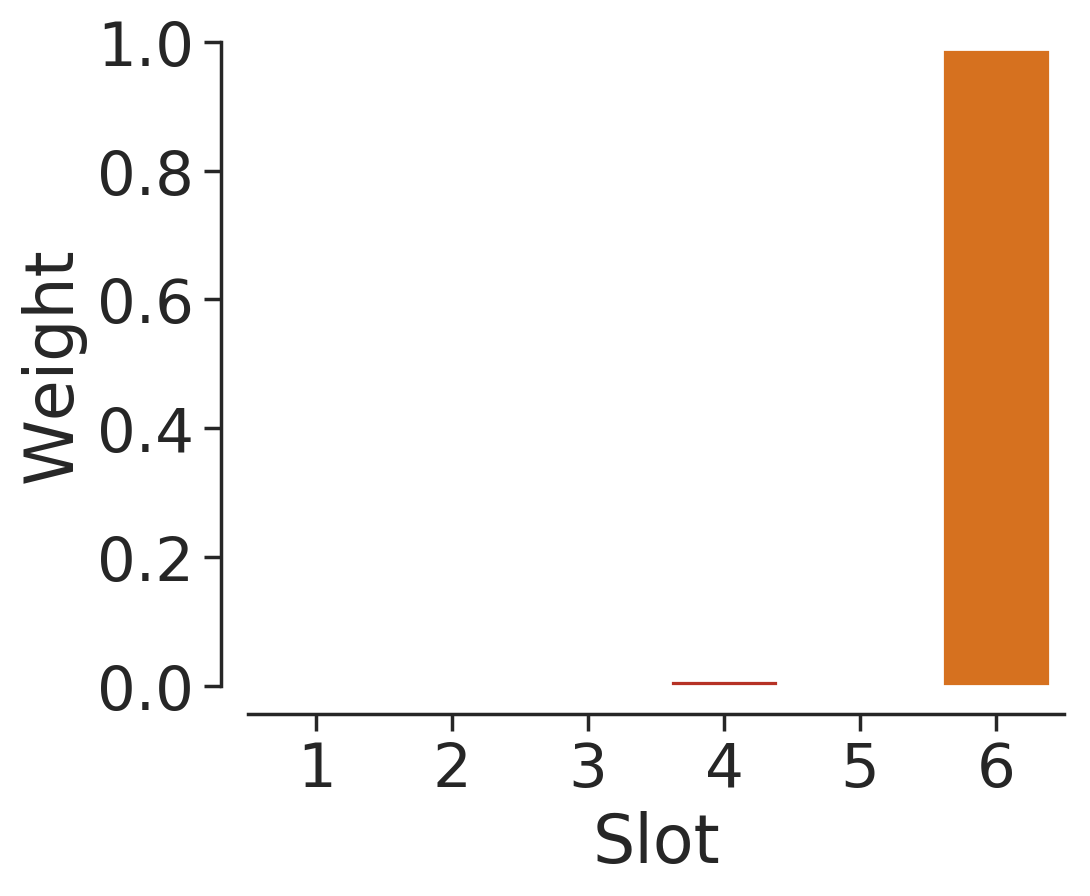} &
        \includegraphics[width=\csize\textwidth]{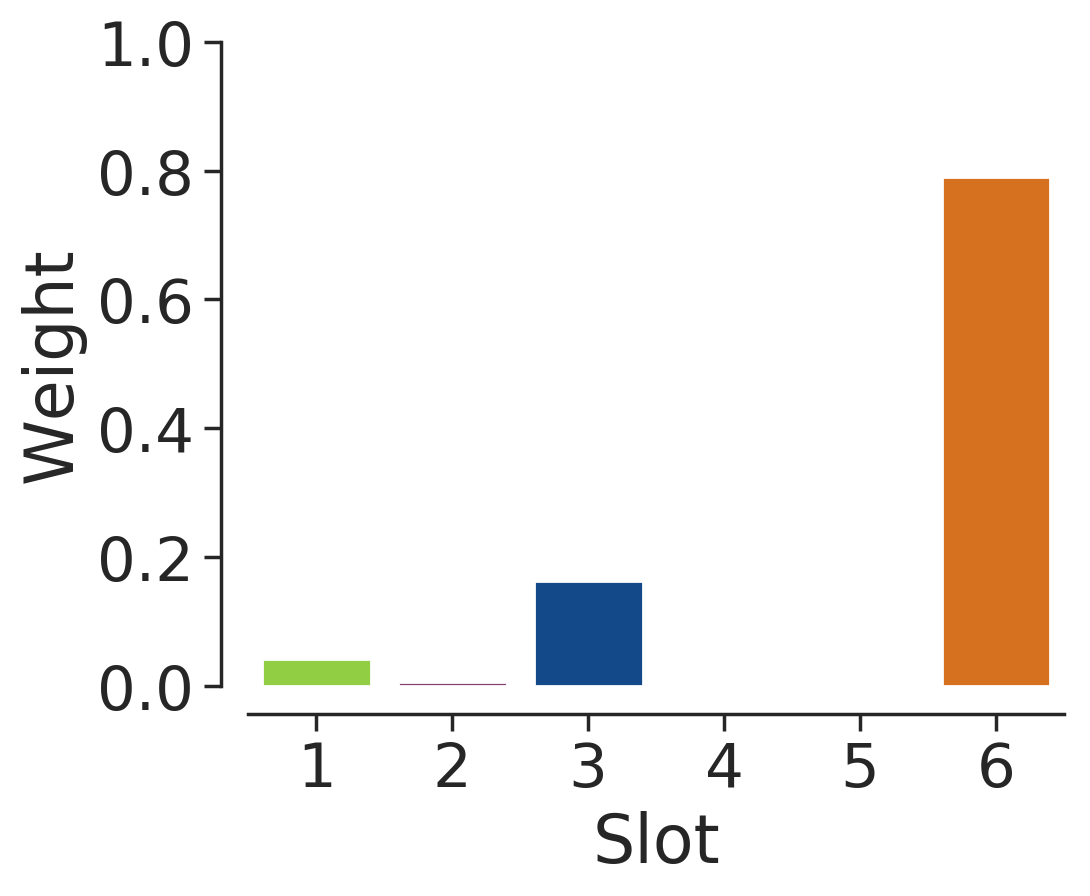} &
        \includegraphics[width=\csize\textwidth]{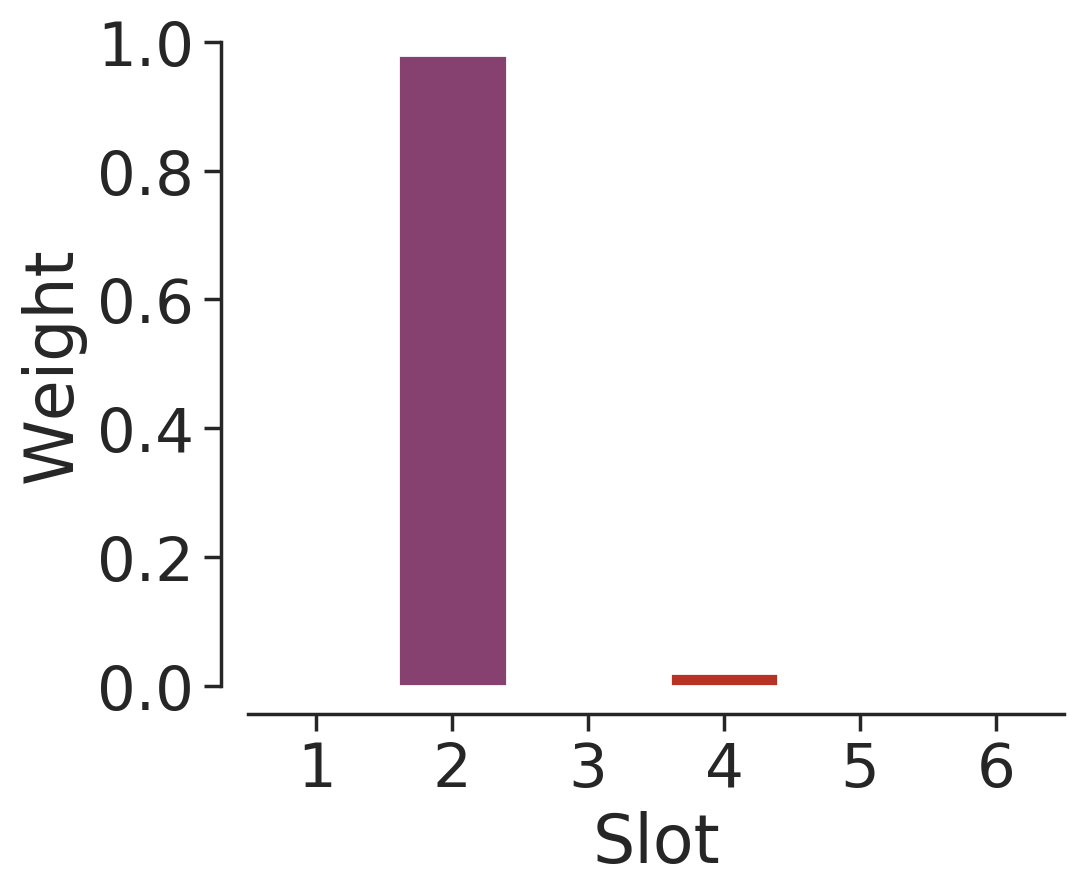} &
        \includegraphics[width=\csize\textwidth]{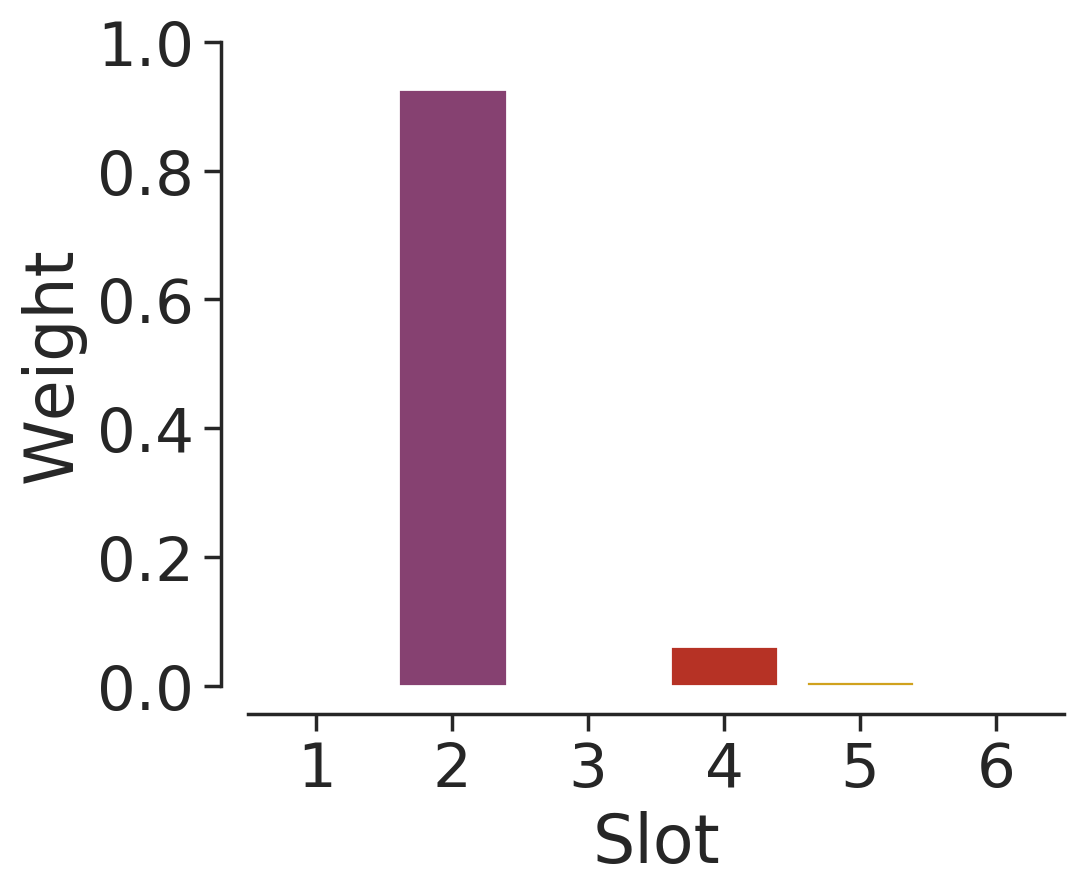} & 
        \includegraphics[width=\csize\textwidth]{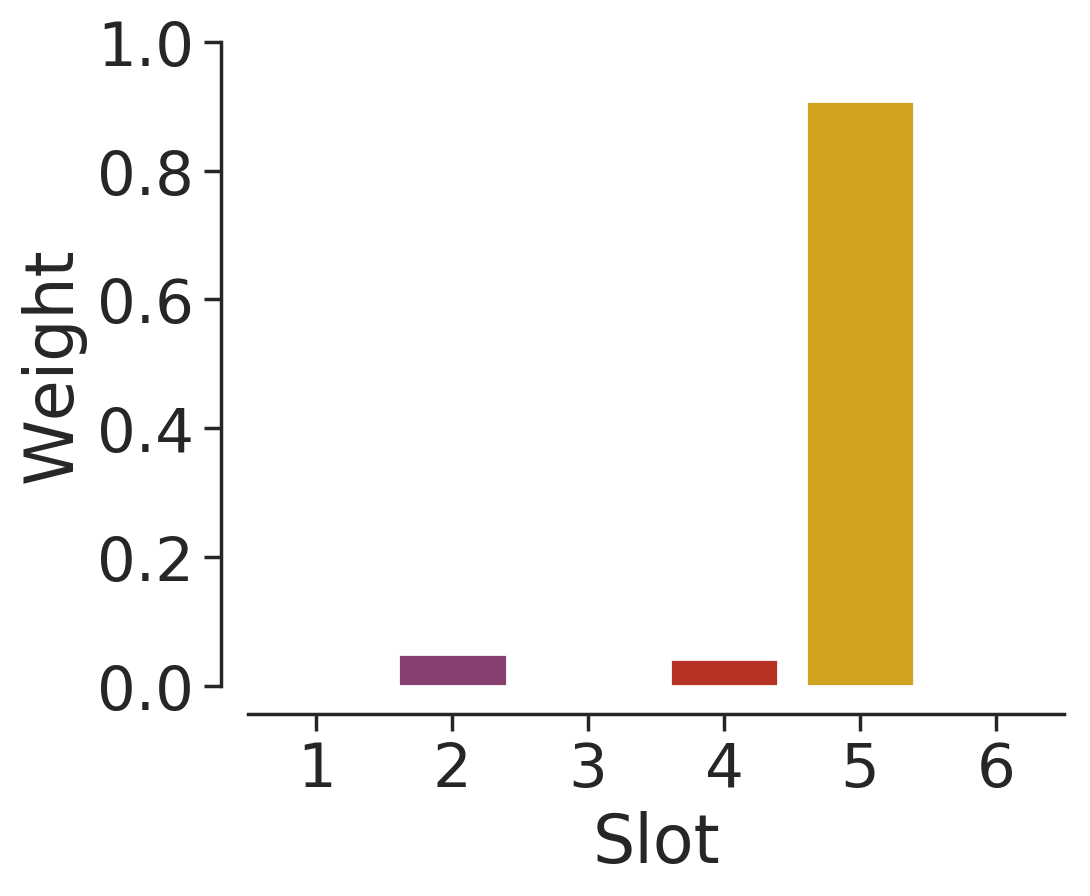} & 
        \includegraphics[width=\csize\textwidth]{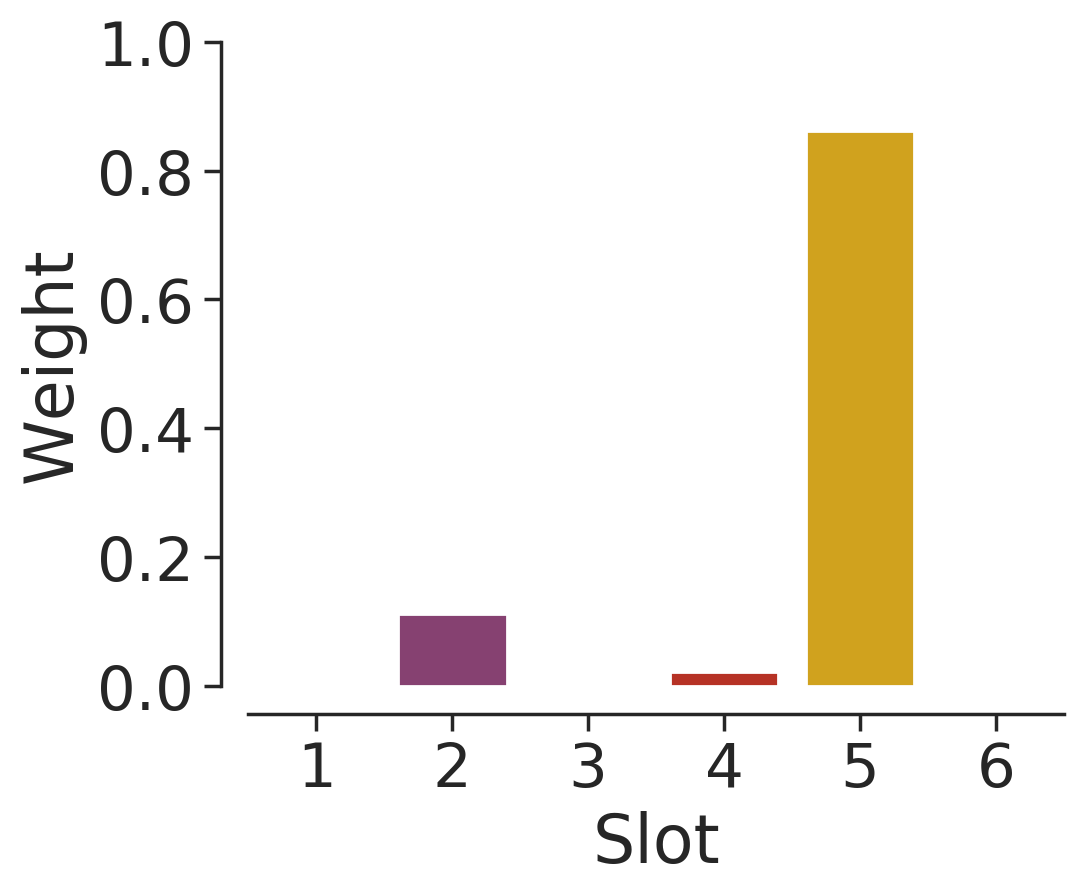} & 
         & \\
        \midrule
        3 towers &
        \includegraphics[width=\csize\textwidth]{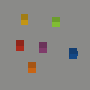} & 
        \includegraphics[width=\csize\textwidth]{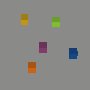} & \includegraphics[width=\csize\textwidth]{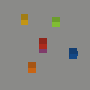} & \includegraphics[width=\csize\textwidth]{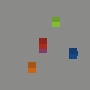} &
        \includegraphics[width=\csize\textwidth]{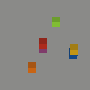} &
        \includegraphics[width=\csize\textwidth]{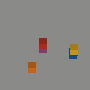} &
        \includegraphics[width=\csize\textwidth]{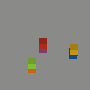} & 
         & & & \\
         & &
        \includegraphics[width=\csize\textwidth]{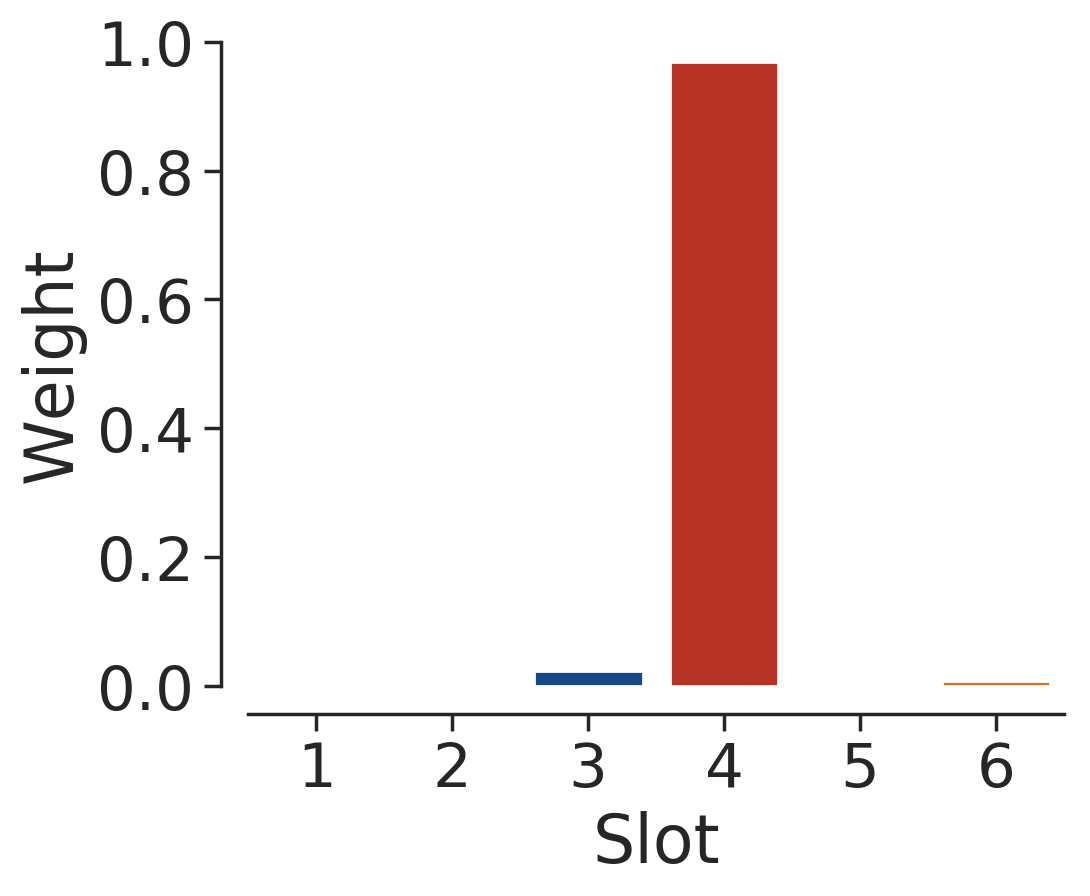} & \includegraphics[width=\csize\textwidth]{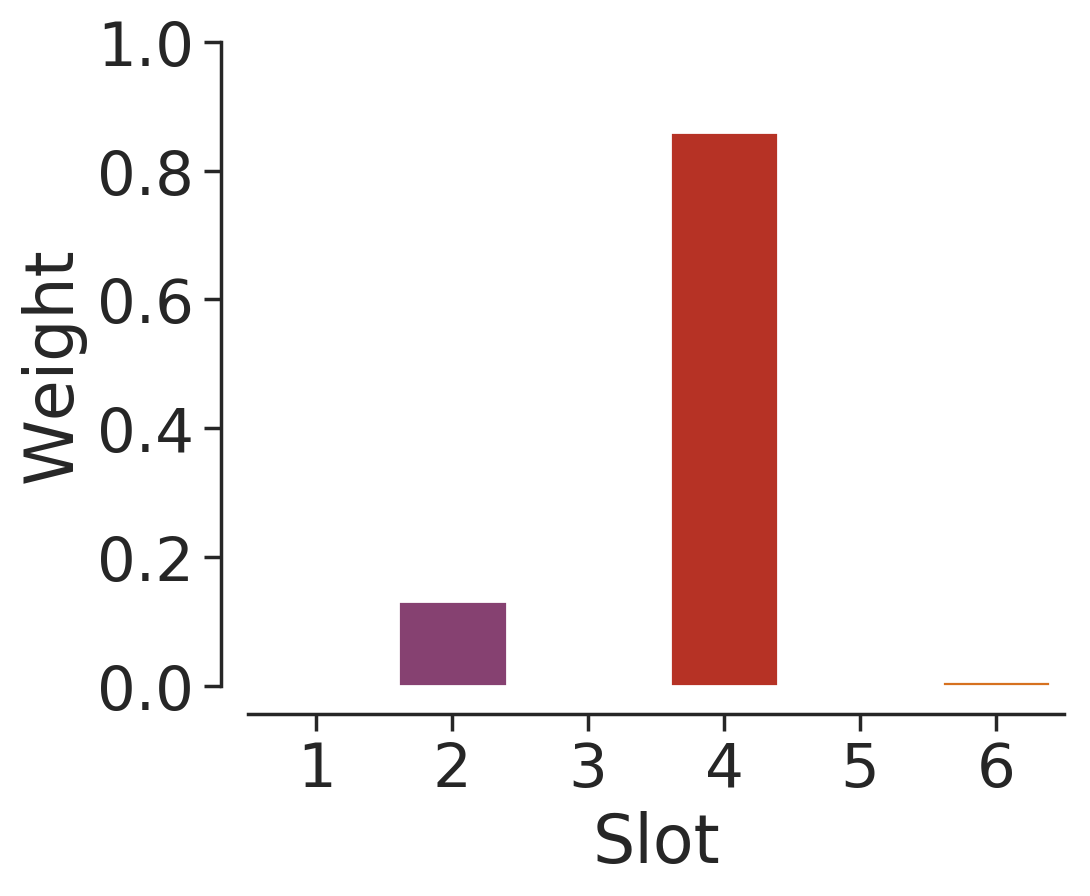} & \includegraphics[width=\csize\textwidth]{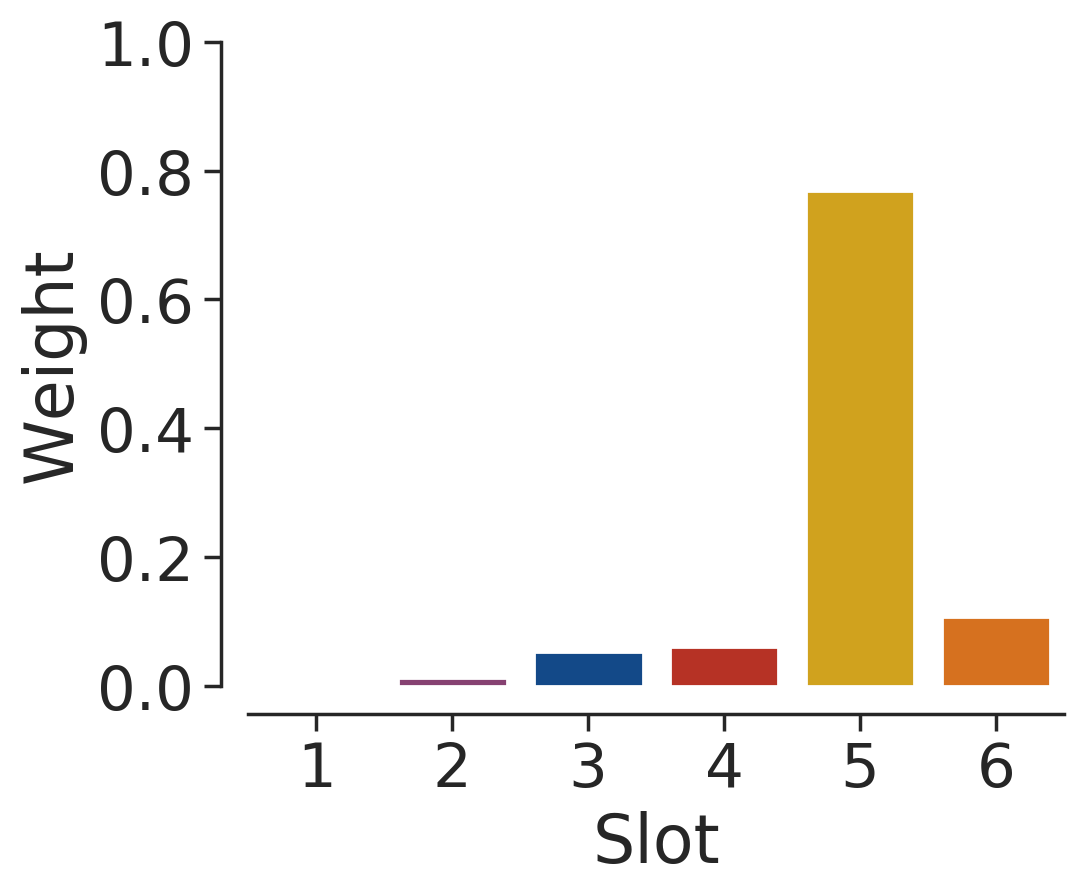} &
        \includegraphics[width=\csize\textwidth]{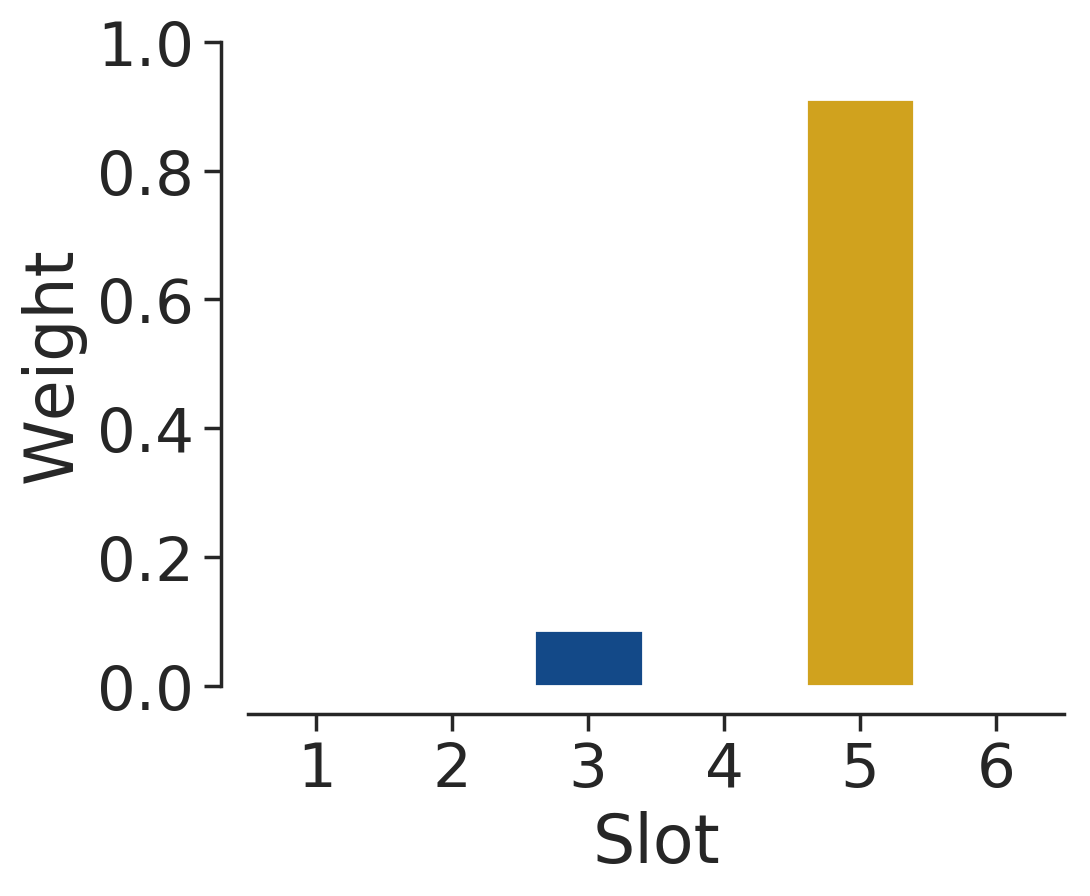} &
        \includegraphics[width=\csize\textwidth]{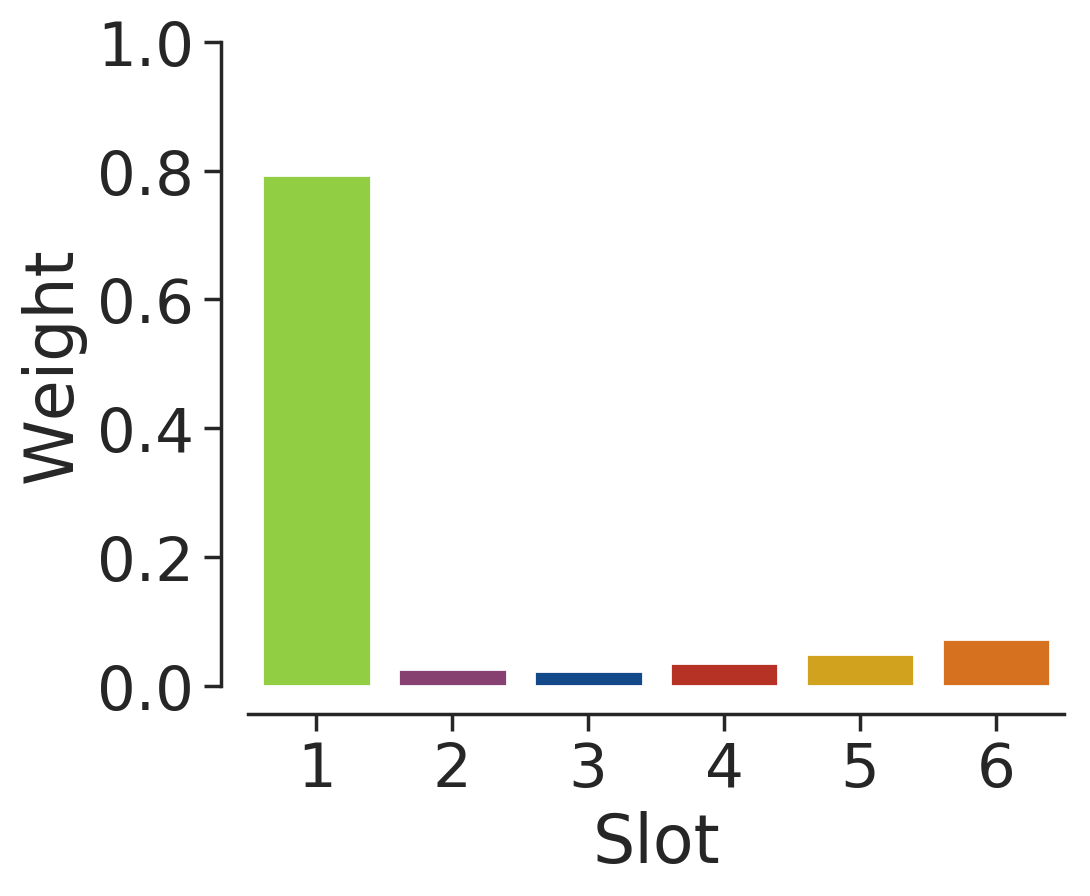} &
        \includegraphics[width=\csize\textwidth]{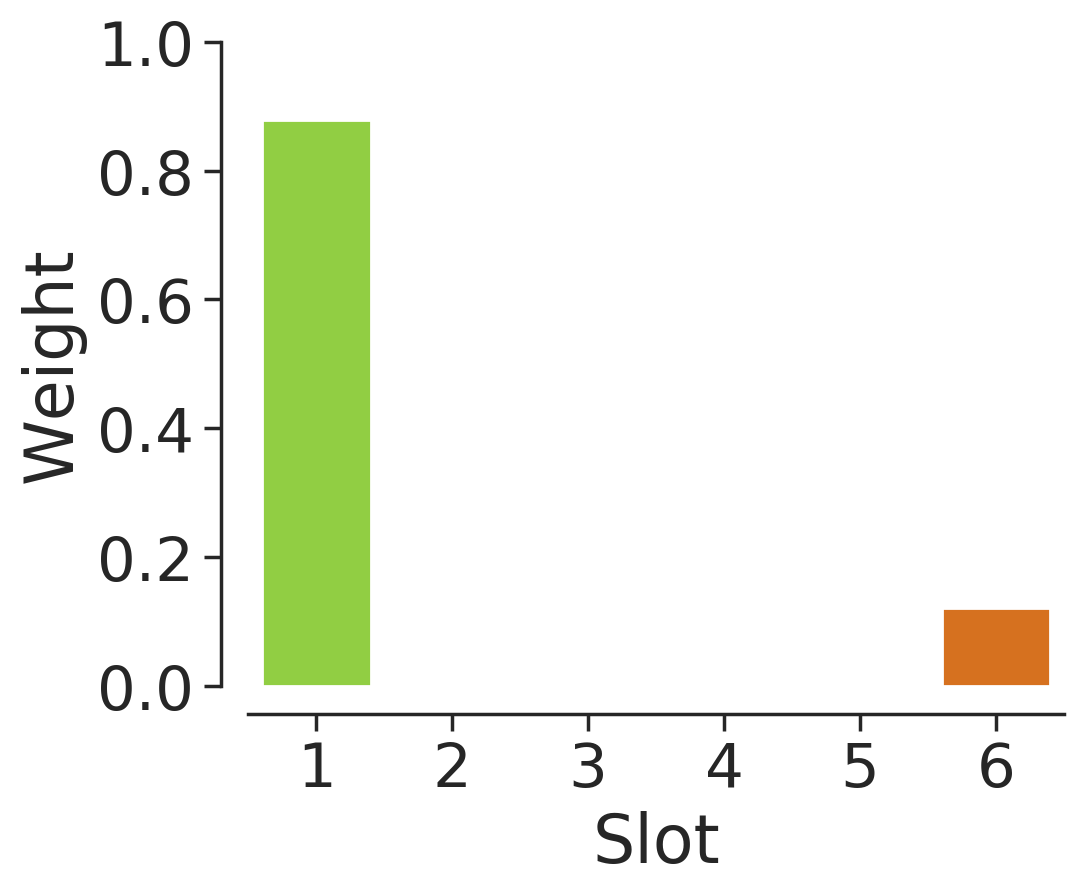} & 
         & & & \\
        \bottomrule
    \end{tabular}
    \caption{Visualization of attention weights for the four zero-shot transfer tasks in our pick-and-place experiment. Individual bars are connected to objects in the images by their colors.}
    \label{tab:fwm_attention}
\end{table}
\end{flushleft}

\end{document}